\documentclass[sn-chicago]{sn-jnl}

\usepackage{natbib}
\usepackage{graphicx}%
\usepackage{multirow}%
\usepackage{amsmath,amssymb,amsfonts}%
\usepackage{amsthm}%
\usepackage{mathrsfs}%
\usepackage[title]{appendix}%
\usepackage{xcolor}%
\usepackage{textcomp}%
\usepackage{manyfoot}%
\usepackage{booktabs}%
\usepackage{algorithm}%
\usepackage{algorithmicx}%
\usepackage{listings}%

\usepackage[utf8]{inputenc}

\usepackage{hyperref}       
\usepackage{booktabs}       
\usepackage{amsfonts}       
\usepackage{nicefrac}       

\usepackage{xcolor}         
\usepackage{wrapfig}        

\usepackage{comment}

\usepackage{amssymb}
\usepackage{amsthm}

\usepackage{blindtext}
\usepackage{booktabs}
\usepackage{varwidth}
\usepackage{caption}

\usepackage{graphicx}

\usepackage{paralist}
\usepackage{mathrsfs}
\usepackage{upgreek}
\usepackage{newtxmath} 
\usepackage{cleveref}
\usepackage{multirow} 
\usepackage{enumitem}

\newtheorem{proposition}{Proposition}

\newtheorem{remark}{Remark}

\usepackage{algorithm}
\usepackage[noend]{algpseudocode}

\usepackage{tikz}
\usetikzlibrary{shapes, arrows}
\usepackage{capt-of}

\usetikzlibrary{arrows.meta}
\tikzset{>={Latex[width=2mm,length=2mm]}}

\usepackage{url} 

\defcitealias{rosenbaum1983central}{RR83}

\theoremstyle{thmstyleone}%

\usepackage{filecontents}
\usepackage{multibib}
\newcites{supp}{Supplementary References}

\let\cline\cmidrule

\begin{document}

\title[Article Title]{Stratified Learning: A General-Purpose Statistical Method for Improved Learning under Covariate Shift}

\author*[1]{\fnm{Maximilian} \sur{Autenrieth}}\email{m.autenrieth19@imperial.ac.uk}

\author[1]{\fnm{David  A.} \sur{van Dyk}}\email{d.van-dyk@imperial.ac.uk}

\author[1,2]{\fnm{Roberto} \sur{Trotta}}\email{rtrotta@sissa.it}

\author[3]{\fnm{David C.} \sur{Stenning}}\email{dstennin@sfu.ca}

\affil*[1]{\orgdiv{Department of Mathematics}, \orgname{Imperial College London}}
\affil[2]{\orgdiv{Department of Theoretical and Scientific Data Science}, \orgname{SISSA (Trieste)}}
\affil[3]{\orgdiv{ Department of Statistics and Actuarial Science}, \orgname{Simon Fraser University}}

\abstract{
We propose a simple, statistically principled, and theoretically justified method to improve supervised learning when the training set is not representative, a situation  
known as covariate shift. We build upon a well-established methodology in causal inference, and show that the effects of covariate shift can be reduced or eliminated by conditioning on propensity scores. In practice, this is achieved by fitting learners within 
strata constructed by partitioning the data based on the estimated propensity scores, leading to approximately balanced covariates and much-improved target prediction.
We demonstrate the effectiveness of our general-purpose method on two contemporary research questions in cosmology, outperforming state-of-the-art importance weighting methods.
We obtain the best reported AUC (0.958) on the updated “Supernovae photometric classification challenge”, and we improve upon existing 
conditional density estimation of galaxy redshift from Sloan Data Sky Survey (SDSS) data.
}


\maketitle

\section{Introduction}\label{sec:introduction}


In supervised learning, when the labeled training (source) data do not accurately represent the unlabeled (target) data, 
learning models may not generalize well, leading to unreliable target output predictions. 
Domain adaptation methods aim to obtain accurate target predictions under such domain shift \citep{pan2009survey}. 
Domain adaptation scenarios 
are widespread, 
and a variety of methods to tackle them have been proposed. Following \cite{kouw2019review}, they can be organized into three categories: 
feature-based methods \citep[e.g.,][]{pan2010domain}; 
inference-based methods  \citep[e.g.,][]{liu2014robust}; 
and sample-based 
approaches, with a focus on importance weighting \citep[e.g.,][]{shimodaira2000improving,sugiyama2008direct,cortes2010learning}, mainly in the covariate shift framework, which is our focus. 
(Section~\ref{section:supplement_bibnote} of the Supplement includes additional background.)

\paragraph{Covariate Shift:}
Adapting a model trained on unrepresentative source data to accurately predict target data labels requires information about the target data distribution.
We consider covariate shift, a common case of domain shift.
Specifically, let $\mathcal{X} \subset \mathbb{R}^F$, $F> 0$, be the feature space and $\mathcal{Y}$ the label space with $K>1$ classes, or a subset of $\mathbb{R}^K$ in multivariate regression with $K$ dependent variables. Different domains are defined as different joint distributions $p(x,y)$ over the same feature-label space $\mathcal{X} \times  \mathcal{Y}$ \citep{kouw2019review}. 
We consider a transductive, unsupervised domain adaptation case, where source data $D_S= \{(x_S^{(i)}, y_S^{(i)} )\}_{i = 1}^{n_s}$ with $n_s$ labelled samples from the joint distribution $p_S$ (source domain $\mathcal{D}_S$) available, as well as target data $D_T= \{x_T^{(i)}\}_{i = 1}^{n_t}$ with $n_t$ unlabelled samples from the joint distribution $p_T$ (target domain $\mathcal{D}_T$). To avoid the trivial 
case, we assume that $p_S(x,y) \neq p_T(x,y)$. For ease of notation, we implicitly condition on an indicator variable $S$, with $p_S(x,y) := p(x,y | s= 1)$ representing source data (analogously $p_T(x,y) := p(x,y | s= 0)$ for target data). 
Covariate shift is a particular domain adaptation problem, where the conditional distribution of the output variable given the predictive covariates is the same for source and target data, but the distribution of source and target covariates differ, i.e., 
$p_S(y|x) = p_T(y|x)$ but $p_S(x) \neq p_T(x)$ \citep{moreno2012unifying}.

Covariate shift commonly occurs if training samples are not selected randomly, but biased in terms of certain covariates. 
For instance, brighter astronomical objects are more likely to be better 
observed and therefore included in the training set \citep{lima2008estimating,revsbech2018staccato}. 
Selection bias has been widely studied in the statistical literature \citep{heckman1979sample,little2019statistical}, 
e.g.,  when estimating treatment effects via observational studies. 
The transfer of causal inference techniques to domain adaptation 
has received more attention in recent years 
-- both fields share the goal of obtaining accurate estimators under distribution shift 
\citep{magliacane2018domain}.
We build on this work in the present paper.

\paragraph{Propensity Scores:}
The introduction of propensity scores \citep[][hereafter  \citetalias{rosenbaum1983central}]{rosenbaum1983central}  
was groundbreaking in causal inference for obtaining unbiased treatment effect estimates from confounded, observational data. \citetalias{rosenbaum1983central} define the propensity score as the probability of treatment assignment given the observed covariates. They show that, under certain assumptions, conditional on the propensity scores the treatment and control group have balanced covariates, which allows unbiased  
treatment effect estimation. 
Four main methods are used to condition on the propensity scores: inverse probability of treatment weighting (IPTW), using the propensity score for covariate adjustment, matching, and stratification on the propensity scores 
(\citetalias{rosenbaum1983central}; \citealp{rosenbaum1984reducing}). 
Extensive work has been done on best practice and generalization of propensity scores in causal inference, too much to list here, and we refer to  \cite{imbens2015causal} for an overview. 
Propensity scores have also found wide application to related areas, 
such as classification with imbalanced classes \citep{rivera2014oups}, 
or fairness-aware machine learning \citep{calders2013controlling}, among others. 
In the covariate shift framework, estimated propensity scores are used only implicitly for importance weighting \citep[e.g.,][]{zadrozny2004learning,kanamori2009least} 
analogously to IPTW, and for matching to obtain validation data \citep{chan2020unlabelled}. There has been no effort, however, to transfer the general methodology.

In the causal inference literature, there is an ongoing debate between the use of weighted estimators and stratification/matching \citep[e.g.,][]{rubin2004principles,lunceford2004stratification,austin2017performance}. 
While, under correct specification of the propensity score model, weighting leads to consistent estimation of treatment effects, this might not hold for stratification due to potential residual confounding within strata \citep{lunceford2004stratification}. However, the bias introduced by stratified estimators is traded with reduced variance compared to weighted estimators \citep{lunceford2004stratification}. 
In addition, estimates of the propensity score are less variable 
than estimates of their reciprocal, or similarly a density ratio (to form weights). 
A small change in an estimated propensity score 
that is near zero can lead to a large difference in the computed inverse-propensity weight, causing massive variance in the estimates based on the weights \citep{lunceford2004stratification,austin2017performance}.

\paragraph{Contribution:}
We propose stratified learning (\textit{StratLearn}), a simple and statistically principled framework to improve 
learning under covariate shift, based on propensity score stratification.
We show that, theoretically, conditioning on estimated propensity scores eliminates the effects of covariate shift. 
In practice, we partition (stratify) the (source and target) data into subgroups based on the estimated propensity scores, improving covariate balance 
within strata. We show that 
supervised learning models can then be optimized within 
source strata without further adjustment for covariate shift, leading to reduced bias in the predictions for each stratum. \textit{StratLearn} is general-purpose, meaning it is in principle applicable to any supervised regression or classification task under any model.
We provide theoretical evidence for the effectiveness of \textit{StratLearn}, and demonstrate it in a range of low- and high-dimensional applications. 
We show 
that the principled transfer of the propensity score methodology from causal inference  
to the covariate shift framework allows the statistical learning community to employ hard-won practical advice from causal inference, 
e.g., balance diagnostics, propensity score model assessment/selection, etc.
\citep[e.g.,][]{rosenbaum1984reducing,imai2004causal,Austin2007,Pirracchio2014}. 
We stratify to condition on propensity scores instead of using importance weighting to avoid the massive variance associated with the latter.
\textit{StratLearn} (stratification) does not use individually estimated propensity score values except to form strata,  leading to a more robust method \citep{rubin2004principles}, as demonstrated in our numerical studies.

This article is organized as follows. In Section~\ref{sec:preliminaries}, we formally introduce the target risk minimization task, and summarize related literature on covariate shift, particularly on importance weighting methods. We conclude Section~\ref{sec:preliminaries} with an overview of propensity scores in causal inference, before developing our new methodology in Section~\ref{section:methodology}.
In Section~\ref{section:demonstration}, we numerically evaluate our method. 
In Section~\ref{section:Classification_SPCC}, we apply \textit{StratLearn} to Type~Ia supernovae (SNIa) classification, which has attracted broad interest in recent years \citep{kessler2010supernova,boone2019avocado}. 
Improving upon \cite{revsbech2018staccato}, \textit{StratLearn} obtains the best reported AUC\footnote{The AUC is the area under the Receiver Operator Characteristic (ROC) curve, obtained by plotting classifier efficiency against the false positive rate for different classification thresholds (between $[0,1]$). 
} (0.958) on the updated “Supernovae photometric classification challenge” (SPCC) \citep{kessler2010results}. 
In Section~\ref{section:photoZ}, we improve upon 
non-parametric full conditional density estimation of galaxy photometric redshift (i.e., photo-$z$) \citep{izbicki2017photo}, a key quantity in 
cosmology. 
Supplemental Materials (numbers appearing with a prepended `S') 
provide details complementing the numerical results in  Section~\ref{section:demonstration} (Sections~\ref{section:supplement_upSPCC} and \ref{section:supplement_photoZ}), 
and additional numerical evidence using data from the UCI repository \citep{Dua:2019} in Section~\ref{section:supplement_UCI}, and a variation of the SPCC data \citep{kessler2010supernova} in Section \ref{section:supplement_originalSPCC}. 
For additional background, we provide a bibliographic note (Section~\ref{section:supplement_bibnote}), details on related methods (Section~\ref{section:supplement_comparison_models}), data and software used in this paper (Section~\ref{section:supplement_data_software}), as well as complementary details of our proposed methodology \textit{StratLearn} (Section~\ref{section:supplement_methodology}). 
\textit{StratLearn} is computationally efficient, easy to implement and readily adaptable to various applications. 
Our investigations show that \textit{StratLearn} is competitive with state-of-the-art importance weighting methods 
in lower dimensions
and greatly beneficial 
for higher-dimensional applications.

\section{Preliminaries}
\label{sec:preliminaries}

\subsection{Target Risk Minimization:}\label{subsection:target_risk}
In a supervised learning task, let $f:\mathcal{X} \rightarrow \mathbb{R}^K $ be the training function, 
and $\ell : \mathbb{R}^K \times \mathcal{Y} \rightarrow  [0, \infty )$ be the loss function comparing the output of $f$ with the true outcome $\mathcal{Y}$. 
This describes a general multivariate regression case; in a probabilistic classification task with $K$ classes we usually have $f:\mathcal{X} \rightarrow [0,1]^K$.
The risk function associated to our supervised learning task is
$\mathcal{R} (f) := \mathbb{E}[\ell ( f(x),y) ]$.
We cannot generally compute $\mathcal{R} (f)$, since the exact joint distribution $p(x,y)$ is unknown. However, an approximation of the risk can be obtained by computing the empirical risk by averaging the loss on the training sample. 
The objective is to minimize the target risk 
$\mathcal{R}_T (f) := \mathbb{E}_{p_T(x,y)} [\ell ( f(x),y) ]$, via the labelled source data $D_S$ and unlabelled target data $D_T$. 
Our task is to train a model function $f$ that minimizes $\mathcal{R}_T (f)$, being able to compute only the source loss $\ell ( f(x_S),y_S)$, but not the target loss $\ell ( f(x_T),y_T)$. 
Section~\ref{subsection:weighting} reviews 
importance weighting methods to minimize the target risk under covariate shift.

\subsection{Related Literature -- Importance Weighting:} \label{subsection:weighting}
In an influential work, \cite{shimodaira2000improving} proposes a weighted maximum likelihood estimation (MWLE) and shows that this MWLE converges in probability to the minimizer of the target risk. 
Following \cite{shimodaira2000improving}, assuming that the support of $ p_T(x)$ is contained in  $ p_S(x)$, the expected loss (risk) w.r.t. $\mathcal{D}_T$ equals that w.r.t. $\mathcal{D}_S$ with weights $w(x) := p_T(x)/ p_S(x)$ for the loss incurred by each $x$, 
\begin{equation}\label{form:risk_shimodaira}
    \mathbb{E}_{\mathcal{D}_T} \left[\ell ( f(x),y)\right] =  
    \mathbb{E}_{\mathcal{D}_S} \left[ w(x)   \ell ( f(x),y)\right].
\end{equation}  
In short, the target risk can be minimized by weighting the source domain loss by a ratio of the densities of target and source domain features.
The importance weights $w(x)$ are
paramount 
in the covariate shift literature and several approaches optimize the estimation of the weights. One approach estimates the densities $p_T(x)$ and $ p_S(x)$ separately \citep{shimodaira2000improving}, e.g., through kernel density estimators \citep{sugiyama2005input}. 
Others estimate the density ratio directly, e.g., via Kernel-Mean-Matching \citep{huang2007correcting}, 
Kullback-Leibler importance estimation (KLIEP) \citep{sugiyama2008direct}, 
and variations of unconstrained least-squares importance fitting (uLSIF) \citep{kanamori2009least}. 
Given $w(x)$, \cite{sugiyama2007covariate} propose importance weighted cross-validation (IWCV) 
and show that in theory this can deliver an almost unbiased estimate of the target risk.
\cite{zadrozny2004learning} links covariate shift with selection bias, and shows that
the target risk can be minimised by importance sampling of source domain data, employing the inverse probability of source set assignment for importance weights.
This allows any probabilistic classifier to be used to obtain the weights, e.g., logistic regression \citep{bickel2007dirichlet}.

Although importance weighting in theory enables minimization of the target risk, there are challenges. Based on a measure of domain dissimilarity (e.g.,  R\'enyi divergence), 
\cite{cortes2010learning} show that weighting leads to high generalization upper error bounds, making predictions unreliable, especially with large importance weights.  
In addition, \cite{reddi2015doubly} points out that while weighting can reduce bias, it can also greatly increase variance.  
Unfortunately, with increasing feature space dimension, the variance of the importance-weighted empirical risk estimates may increase sharply \citep{izbicki2014high,stojanov2019low}. This can be partly tackled by dimensionality reduction methods \citep{stojanov2019low}; 
see 
\cite{kouw2019review} for a detailed discussion. 
We address these variance concerns via propensity scores.

\subsection{Related Literature -- Propensity scores in causal inference:}\label{subsection:propensity_scores}
Given a set of observed covariates $X$ and a binary indicator $Z$ for treatment assignment (treatment vs control), 
\citetalias{rosenbaum1983central} introduce the propensity score as
\begin{align}\label{form:propensity_score}
 e(X) := P(Z=1|X) 
\end{align}
and define treatment allocation $Z$ as strongly ignorable, if
    \begin{align}\label{form:stronglyignorable}
    \text{(i) }  \ (Y_1, Y_0) \ \Perp Z |X \qquad \text{  and  } \qquad \text{(ii) } \ 0 < e(X) < 1. 
\end{align} 
Condition (i) means that treatment assignment $Z$ is conditionally 
independent of the potential outcome $(Y_1,\ Y_0)$, given the observed covariates. The potential outcomes are the possible outcomes for an object, depending on its treatment status, and
at most one is observed, 
(e.g., for a treated object the observed outcome is $Y = Y_1$). In practice, condition (i) means that no confounders (covariates that are associated with the treatment and outcome) are unmeasured. 
\citetalias{rosenbaum1983central} show that if (\ref{form:stronglyignorable}) holds, the propensity score is a balancing score. That is, given the propensity score, the  distribution of the covariates in treatment and control are the same, i.e., $p(X| e(X), Z = 1) = p(X|e(x), Z = 0)$. Thus, conditional on the propensity score, unbiased average treatment effect estimates can be obtained, i.e., $\mathbb{E}[Y_1 | e(x), Z = 1] - \mathbb{E}[Y_0 | e(x), Z = 0]  = \mathbb{E}[Y_1 - Y_0 | e(x) ]  $. 
In practice, conditioning on the estimated (rather than true) propensity score can achieve better empirical balance as this corrects for statistical fluctuations in the sample as well  (\citetalias{rosenbaum1983central}; \citealp{hirano2003efficient}).

Below, we show how the propensity score methodology can be transferred to the covariate shift framework, for target risk minimization in supervised learning tasks.

\section{A New Method: StratLearn}\label{section:methodology}

\subsection{StratLearn -- Methodology} 

In the covariate shift framework, we 
define the propensity score to be the probability that object $i $ is in the source data, given its observed covariates, i.e., 
\begin{align}\label{form:PS_covariate_shift}
e(x_i) := P(s_i = 1|x_i), \text{ with } 0 < e(x_i) < 1. 
\end{align}

\begin{proposition}[Learning conditional on the propensity score]\label{prop:cond_PS}
If $p_S(x,y)$ and $p_T(x,y)$
satisfy the covariate shift definition and $0<e(x)<1$, then it holds that
\begin{equation}
p_T(x,y | e(x)) = p_S(x,y | e(x)).
\end{equation}
That is, conditional on $e(x)$ the joint source and target distributions are the same, eliminating covariate shift. It follows, for any loss function $\ell = \ell (f(x),y)$,
\begin{align}\label{form:exp_cond_ps}
\mathcal \mathbb{E}_{ p_T(x,y |e(x))} [\ell ( f(x),y) ]  =  \mathcal \mathbb{E}_{ p_S(x,y |e(x))} [\ell ( f(x),y) ]. 
\end{align} 
\end{proposition}
\noindent
\Cref{prop:cond_PS} is verified in Section~\ref{section:supplement_methodology}. 
Note that its condition, $0 < e(x) < 1 $, is no stronger than the 
conditions required for 
(\ref{form:risk_shimodaira}). 
The support of $ p_T(x)$ being contained in  $ p_S(x)$ implies $0 < e(x),$ and $e(x) = 1$ implies 
$p_T(x)=0$, in which case the importance weight $w(x)=0$, which is equivalent to discarding the sample.

With Proposition~\ref{prop:cond_PS}, we extend the basic causal inference theory
to use propensity scores in the covariate shift framework.
Conditioning on estimated propensity scores enables statistically principled minimization of the target risk based on source data.
According to Proposition~\ref{prop:cond_PS}, if we were to condition on any single value of the propensity score, the distribution of $x$ and $y$ in the source and target domains would be identical and we could minimize their target risk using the source data alone. Because sample sizes with identical propensity scores are too small in practice for model fitting, we employ an approximation.

\textit{StratLearn} takes advantage of Proposition~\ref{prop:cond_PS} via propensity score stratification;
source data $D_S$ and target data $D_T$ are divided into $k$ non-overlapping subgroups (strata) based on quantiles of the estimated propensity scores.
More precisely, letting $q_j$ be the $j$'th k-quantile of $\{e(x_i) : x_i \in (x_S \cup x_T) \}$, 
for $j \in 1,\dots , k,$  we divide $D_S$ and $D_T$ into 
\begin{equation}\label{form:PS_strata}
      \kern -6pt
      D_{Sj}^{(k)} = \{(x,y) \in D_S {\,:\,}   q_{k-j}  <  e(x) \leq q_{k-j+1} \} \text{ and } 
      D_{Tj}^{(k)} = \{x \in D_T{\,:\,}   q_{k-j}  <  e(x) \leq q_{k-j+1} \},
\end{equation}
where  $q_0 = 0$ and $q_k = 1.$ By Proposition~\ref{prop:cond_PS}, within strata,
\begin{equation}\label{form:stratified_joint_dist}
  p_{T_j}(y,x)  \approx  p_{S_j}(y,x), \text{ for } j \in 1,\dots , k,
\end{equation}
where $S_j$ indicates conditioning on assignment to the $j$'th source stratum (analogously for target $T_j$). It follows that for $j \in 1,\dots , k$, 
\begin{align}\label{form:strata_risk}
\mathbb{E}_{ p_{T_j}(x,y)} [\ell ( f(x),y) ] \approx  \mathbb{E}_{p_{S_j}(x,y)} [\ell ( f(x),y) ].
\end{align}
Thus, we can minimize the target risk within strata by minimizing the source risk within strata. In this way, we reduce the covariate shift problem to non-overlapping 
subgroups where the source and target domain are approximately the same, which in principle allows us to fit any supervised learner to $D_{S_j}$ to predict the target objects in $D_{T_j}$. Figure~\ref{figure:flow_chart} presents a flow chart illustrating the steps of our proposed $\textit{StratLearn}$ methodology.


\tikzstyle{terminator} = [rectangle, draw, text centered, minimum height=2em, text width=20em]
\tikzstyle{block} = [rectangle, draw, text centered, text width=20em, minimum height=2em]
\tikzstyle{block_m} = [rectangle, draw, text centered, text width=20em, minimum height=2em] 
\tikzstyle{outer_box} = [rectangle,  draw, text centered, text width=28em, minimum height=27em, ultra thick] 
\tikzstyle{connector} = [draw, -latex']
\tikzstyle{decision} = [diamond, draw, text badly centered, minimum height=2.2em, inner sep=0pt, text width=4.3em]
\tikzstyle{descr} = [fill=black, inner sep=-0pt]

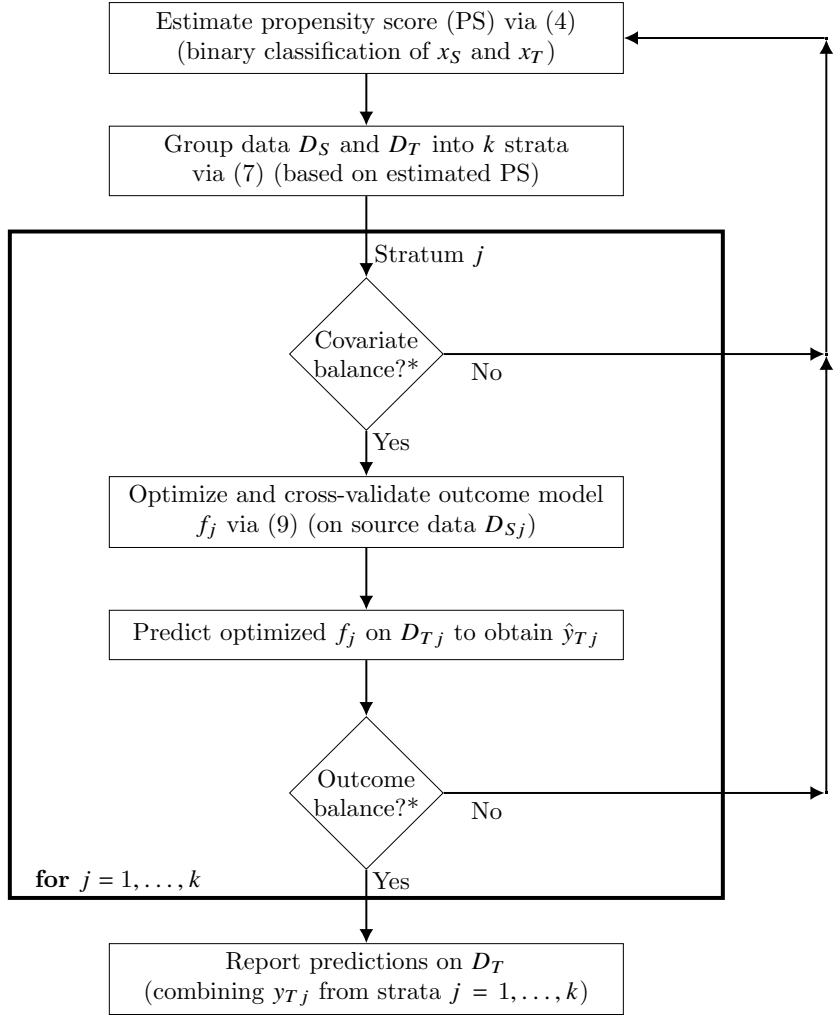
\begin{figure}
\begin{center}
\resizebox{11cm}{!}{%
\begin{tikzpicture}
    \node [terminator] at (0,0) (PS) 
    {Estimate propensity score (PS) via (\ref{form:PS_covariate_shift}) \\
    (binary classification of $x_S$ and $x_T$)};
    \node [descr] at (6.5,0) (PS_connect) {};
    \node [block] at (0,-1.75) (Grouping) 
    {Group data $D_S$ and $D_T$ into $k$ strata \\ via (\ref{form:PS_strata})
    (based on estimated PS)};
    \node [outer_box] at (0,-7.5) (loop_box) {};
    \node [decision] at (0,-4.5) (Cov_balance) 
    {Covariate balance?*};
    \node [descr] at (6.5,-4.5) (Cov_balance_connect) {};
    \node [block_m] at (0,-6.75) (Optimize) 
    {Optimize and cross-validate outcome model \\ $f_j$ via (\ref{form:strata_risk}) 
    (on source data $D_{Sj}$)};
    \node [block] at (0,-8.5) (Predict) 
    {Predict optimized $f_j$ on $D_{Tj}$ to obtain $\hat{y}_{Tj}$};
    \node [decision] at (0,-10.75) (Out_balance) 
    {Outcome balance?*};
    \node [descr] at (6.5,-10.75) (Out_balance_connect) {};
    \node [terminator] at (0,-13.4) (End)
    {Report predictions on $D_T$ \\
    (combining $y_{Tj}$ from strata $j=1,\dots,k$)};    
    \draw[->, thick] (PS) node[]{} -- (Grouping) node[]{};
    \draw[->, thick] (Grouping) node[]{} -- (Cov_balance) node[]{};
    \draw[->, thick] (Cov_balance)  node[]{} -- (Optimize) node[]{};
    \draw[->, thick] (Optimize) node[]{} -- (Predict) node[]{};
    \draw[->, thick] (Predict) node[]{} -- (Out_balance) node[]{};
    \draw[->, thick] (Out_balance) node[]{} -- (End) node[]{};
    \draw[->, thick] (Out_balance_connect) node[]{} -- (Cov_balance_connect) node[]{};
    \draw[->, thick] (Cov_balance_connect) node[left]{} -- (PS_connect) node[right]{};
    \draw[->, thick] (Out_balance) node[left]{} -- (Out_balance_connect) node[right]{};
    \draw[->, thick] (Cov_balance) node[left]{} -- (Cov_balance_connect) node[right]{};
    \draw[->, thick] (PS_connect) node[left]{} -- (PS) node[right]{};   
    \node[draw=none] at (0.9, -3.1) (Stratum_j) {Stratum $j$};
    \node[draw=none] at (-3.5, -12.) (for_loop) {\textbf{for $j= 1,\dots,k$} };
    \node[draw=none] at (1.7, -4.75) (no1) {No};
    \node[draw=none] at (0.35, -5.75) (yes1) {Yes};
    \node[draw=none] at (1.7, -11.) (no1) {No};
    \node[draw=none] at (0.35, -12.) (yes1) {Yes};
    \node[draw=none] at (0.35, -14.2) (empty) { };
\end{tikzpicture}%
}
\end{center}
\captionof{figure}{\textit{StratLearn} flow chart. (\text{*}Covariate balance and outcome balance is assessed as described in Section~\ref{section:balance_diagnostics}, with a numerical example given in Section~\ref{section:Classification_SPCC}.)\label{figure:flow_chart} }
\end{figure}



\subsection{StratLearn -- Technical Details}\label{section:technical_details}
In general, any probabilistic classifier could be used to estimate propensity scores (e.g., \cite{Pirracchio2014}). Logistic regression is commonly used 
in causal inference, and we adopt it for the applications in this paper. In practice, the covariate shift assumption, $p_S(y |x) = p_T(y|x)$, requires there be no unobserved confounding covariates. To meet this requirement, we include all potential confounders as main effects. An estimate of the propensity scores in (\ref{form:PS_covariate_shift}) is then obtained by probabilistic classification of source assignment, 
given source data $x_S$ and target data $x_T$. 
Using the estimated propensity scores, the source and target data are grouped into strata, following (\ref{form:PS_strata}). 
We use $k=5$ strata based on empirical evidence provided by \cite{cochran1968effectiveness}, showing that sub-grouping into five strata is enough to remove at least 90\% of the bias for many continuous distributions \citepalias{rosenbaum1983central}.
Given the stratified data, we fit a model $f_j$ to source data $D_{S_j}$, and predict the respective target samples in $D_{T_j}$, for $j \in 1,\dots , k.$ 
Model hyperparameters for $f_j$ can be selected through empirical risk minimization on source data $D_{S_j}$, for instance through cross-validation on $D_{S_j}$. 
The model functions $f_j$ are trained independently and can be computed in parallel to reduce computational time. 
If the source distribution does not cover the target distribution well enough, some of the strata may contain too little source data to reliably train the model. In this case, we add source data from one or more adjacent strata to avoid highly variable predictions. There is a bias-variance trade-off here in that this reduction in variance requires a relaxation of the approximation in (\ref{form:stratified_joint_dist}), which inevitably increases bias somewhat. Although a general and precise criterion for combining the strata is elusive (more complex models require more data and data sets of the same size may be more or less informative for the same model), we illustrate the combination of source strata in 
Section~\ref{section:Classification_SPCC} (and 
Section~\ref{section:supplement_UCI}), where one or more source strata have insufficient data.

\subsection{StratLearn -- Balance Diagnostics}\label{section:balance_diagnostics}

A key advantage of propensity scores derived in causal inference is their covariate balancing score property \citepalias{rosenbaum1983central}, that is, $p_S(x | e(x)) = p_T(x | e(x))$. 
In causal inference, this property is used to verify the propensity score model and/or the choice of covariates, $x$, e.g., by checking that $x$ has the same within-strata distribution in the treatment and control groups. Employing the balancing property in the derivation of Proposition~\ref{prop:cond_PS} allows us to take advantage of such diagnostic tools in our framework. 
We refer to the large literature on this 
\citep[e.g.,][]{rosenbaum1984reducing,imai2004causal,Austin2007,Austin2011}, 
and provide an example of such a balance check in Section~\ref{section:Classification_SPCC} (and in Sections~\ref{section:supplement_originalSPCC} and \ref{section:supplement_UCI}).

In Remark~\ref{remark:potential_outcomes}, we detail how additional structure in the covariate shift setting can be exploited to justify a corollary model diagnostic.

\bigskip
\begin{remark} \label{remark:potential_outcomes}
In the propensity score framework of causal inference \citep{rosenbaum1983central} we have potential outcomes $Y_0$ and $Y_1$. 
In the covariate shift framework, the potential outcomes are identical
($Y_0 \equiv Y_1 $). That is, there is no ``treatment effect'' from being assigned to the source or target set, though only the source data are observed ($Y_1 \equiv Y $). Now, given the propensity score $e(x),$ with $0< e(x)<1,$ and the covariate shift condition $p(y|x, s = 1) = p(y |x, s= 0)$, source data assignment is `\textit{strongly ignorable}' (using the terminology of \citetalias{rosenbaum1983central}). It follows through Theorem 4 in \citetalias{rosenbaum1983central} that, conditional on the propensity score, source and target outcome are the same in expectation.
\end{remark}
\bigskip

In cases where labels are observed for (part of) the target group we can use Remark~\ref{remark:potential_outcomes} as a model diagnostic. Although in practice the labels are mostly unobserved in the target group, they are available in our real-world scientific/experimental settings described in Sections~\ref{section:Classification_SPCC} and \ref{section:photoZ}. In Section~\ref{section:Classification_SPCC} (as well as Sections~  \ref{section:supplement_originalSPCC}, \ref{section:supplement_UCI} and \ref{section:supplement_photoZ}),
we use Remark~\ref{remark:potential_outcomes} to demonstrate a reduction of within-strata covariate shift (i.e., by conditioning on the propensity score). 

We further demonstrate the possibility of similarly using predicted labels instead of actual labels as a model diagnostic. 
While the actual target labels $y_T$ 
are usually not available in real-world data applications, the distribution of the model predicted outcome labels $(\tilde{y} = f(x))$ can be evaluated for source $f(x_S)$ and target $f(x_T)$. With $f$ being a measurable function of the covariates $x$, and 
by employing the balancing property of propensity scores, it holds $p_S(f(x) | e(x)) = p_T(f(x) | e(x))$.  
Consequently, a discrepancy between the distributions of predicted source outcome $f(x_S)$ and predicted target outcome $f(x_T)$ 
is an indication of residual (covariate) shift in the source and target distribution.\footnote{In practice, one has to ensure that $f$ is not overfitted on the available training (source) data sample, a standard check in supervised learning, which can readily be assessed via standard validation tools, such as cross-validation or bootstrapping of source data.}
An advantage of assessing the balance in the predicted outcome $f(x)$ (in addition to covariate balance) is that $f(x)$ is designed to approximate the outcome $y$. Thus, a discrepancy of $f(x_S)$ and $f(x_T)$ indicates an imbalance in strongly predictive covariates, a straightforward sign for remaining, (likely) concerning confounding. We demonstrate the application of balance diagnostics via predicted labels in Section~\ref{section:Classification_SPCC}.


\section{Numerical Demonstrations} \label{section:demonstration}

\subsection{Comparison Methods} 
\label{section:comparison_models}
We compare \textit{StratLearn} to a range of well-established importance weighting methods. 
\begin{itemize}
  \setlength{\itemsep}{1pt}
  \setlength{\parskip}{0pt}
  \setlength{\parsep}{0pt}
    \item KLIEP -- Kullback-Leibler importance 
    estimation procedure 
    \citep{sugiyama2008direct}. 
    \item uLSIF --  Unconstrained 
    least-squares importance fitting 
    \citep{kanamori2009least}.\footnote{
    \href{http://www.ms.k.u-tokyo.ac.jp/sugi/software.html}{KLIEP and uLSIF were implemented with the original author's public domain MATLAB code (link). } }\label{footnote:KLIEP_uLSIF}     
     \item NN -- Several versions of the nearest-neighbor 
     importance weight estimator  \citep{lima2008estimating,kremer2015nearest,loog2012nearest}, varying the 
     number of neighbors. 
    \item IPS -- Importance weight estimation through probabilistic classification of source set assignment \citep{kanamori2009least}. 
\end{itemize}
\bigskip
 
\noindent
In Section~\ref{section:photoZ},  we incorporate the estimated weights as in the corresponding benchmark publication. Following \cite{izbicki2017photo}, the estimated weights are used for loss weighting as in (\ref{form:risk_shimodaira}). In Section~\ref{section:Classification_SPCC}, importance weighting has not previously been applied. We implement 
IWCV, importance sampling, 
and a combination of both, to demonstrate the advantage of \textit{StratLearn} with respect to either; see 
Section~\ref{section:supplement_comparison_models}.


\subsection{Classification -- SNIa Identification} \label{section:Classification_SPCC}

\paragraph{Objective:} 

Type~Ia supernovae (SNIa) are invaluable for the study of the accelerated expansion history of the universe \citep[e.g.,][]{riess1998observational,perlmutter1999measurements}. SNIa are exploding stars that can be seen at large distances, occurring due to a particular physical scenario which causes their intrinsic luminosities to be (nearly) the same. This ``standard candle'' property of SNIa makes it possible to measure their distance, which in turn depends on parameters that describe the expansion of the universe.

\begin{figure}[t!]
\centering
\begin{minipage}{.99\textwidth}
  \centering
  \includegraphics[width=0.99\linewidth]{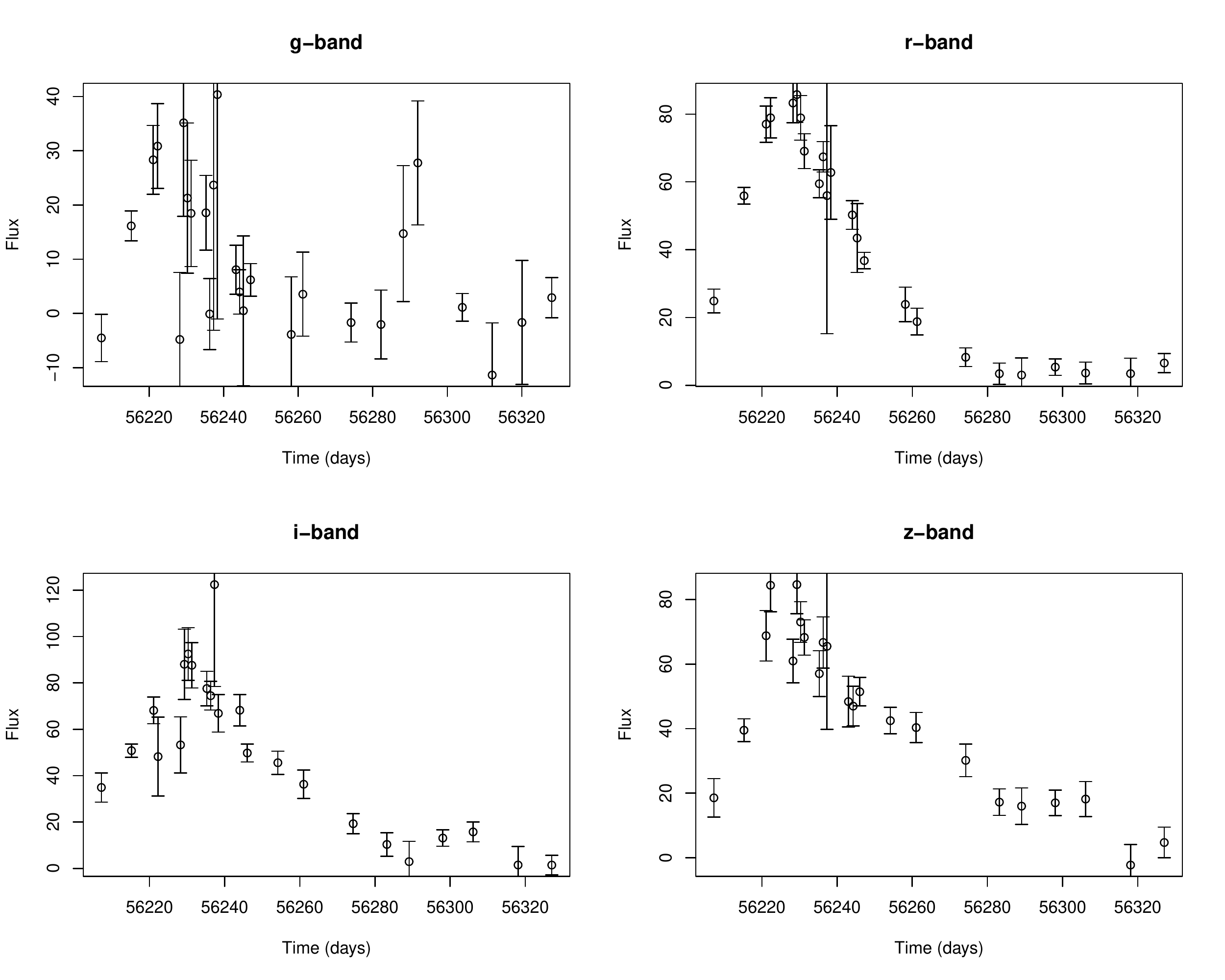}
\end{minipage}
\caption{\baselineskip=15pt 
Example of photometric LC data, including $1\sigma$ error bars, for a typical SNIa (specifically, SN2475 from the updated \cite{kessler2010results} simulated SPCC data).
\label{figure:LC_example}}
\end{figure}

To take advantage of this, reliable identification of SNIa based on photometric light curve (LC) data is a major challenge in modern observational cosmology.
Photometric LC data are easily collectable, consisting of measurements of an astronomical object’s brightness (i.e. flux), filtered through different passbands (wavelength ranges), at a sequence of time points (as illustrated in Fig~\ref{figure:LC_example}). 
Only a small subset of the objects are labelled via expensive 
spectroscopical observations. The labeled source data, $D_S$, are not representative of the photometric target data, $D_T,$ as the selection of spectroscopic source samples is biased towards brighter and bluer objects. 
The automatic classification of 
unlabelled
objects, 
based on biased spectroscopically confirmed source data, is the subject of much research, 
including public classification challenges 
\citep{kessler2010supernova,kessler2019models}.

Leading SNIa classification approaches are based on data augmentation; they  sample synthetic objects from Gaussian process (GP) fits of the LCs to overcome covariate shift \citep{revsbech2018staccato,boone2019avocado}.
The method of \cite{revsbech2018staccato} can be viewed as a prototype of \textit{StratLearn}, as it augments the source data separately in strata based on the estimated propensity scores. However, to optimize data augmentation within strata, \cite{revsbech2018staccato} requires a sub-sample of labeled target data that is unavailable in practice. 
While effective in this particular case, GP data augmentation is not an option in most covariate shift tasks.
We show that \textit{StratLearn} makes augmentation unnecessary. 
We use target prediction AUC to compare performance to published results.


\paragraph{Data and Preprocessing:} We use data from the updated “Supernova photometric classification challenge” (SPCC) \citep{kessler2010results}, containing a total of 21,318 simulated SNIa and of other types (Ib, Ic and II). For each supernova (SN), LCs are given in four color bands, $\{g, r, i, z\}$. The data include a source set $D_S$ of 1102 spectroscopically confirmed SNe with known types and 20,216 SNe with unknown types 
(target set $D_T$). 51\% of the source objects are SNIa, 
while only 23\% of the target date are SNIa, 
a consequence of the strong covariate shift in the data. 

We follow the approach in \cite{revsbech2018staccato}, which was applied to an earlier release of the SPCC data \citep[][discussed in Section~\ref{section:supplement_originalSPCC}]{kessler2010supernova}, to extract a set of features from the LC data that can be used for classification. 
First, a GP with a squared exponential kernel is used to model the LCs. Then, a diffusion map \citep{coifman2006diffusion} (as used in \cite{richards2012semi}) 
is applied, resulting in a vector of 100 similarity measures between the LCs that we use as predictor variables. 
Combining these with redshift (a measure of cosmological distance, defined in Section~\ref{section:photoZ}) and a measure of overall 
brightness, we obtain 102 predictive covariates.

\paragraph{Results:}

\begin{figure}[t!]
\centering
\begin{minipage}{.47\textwidth}
  \centering
    \includegraphics[width=0.95\linewidth]{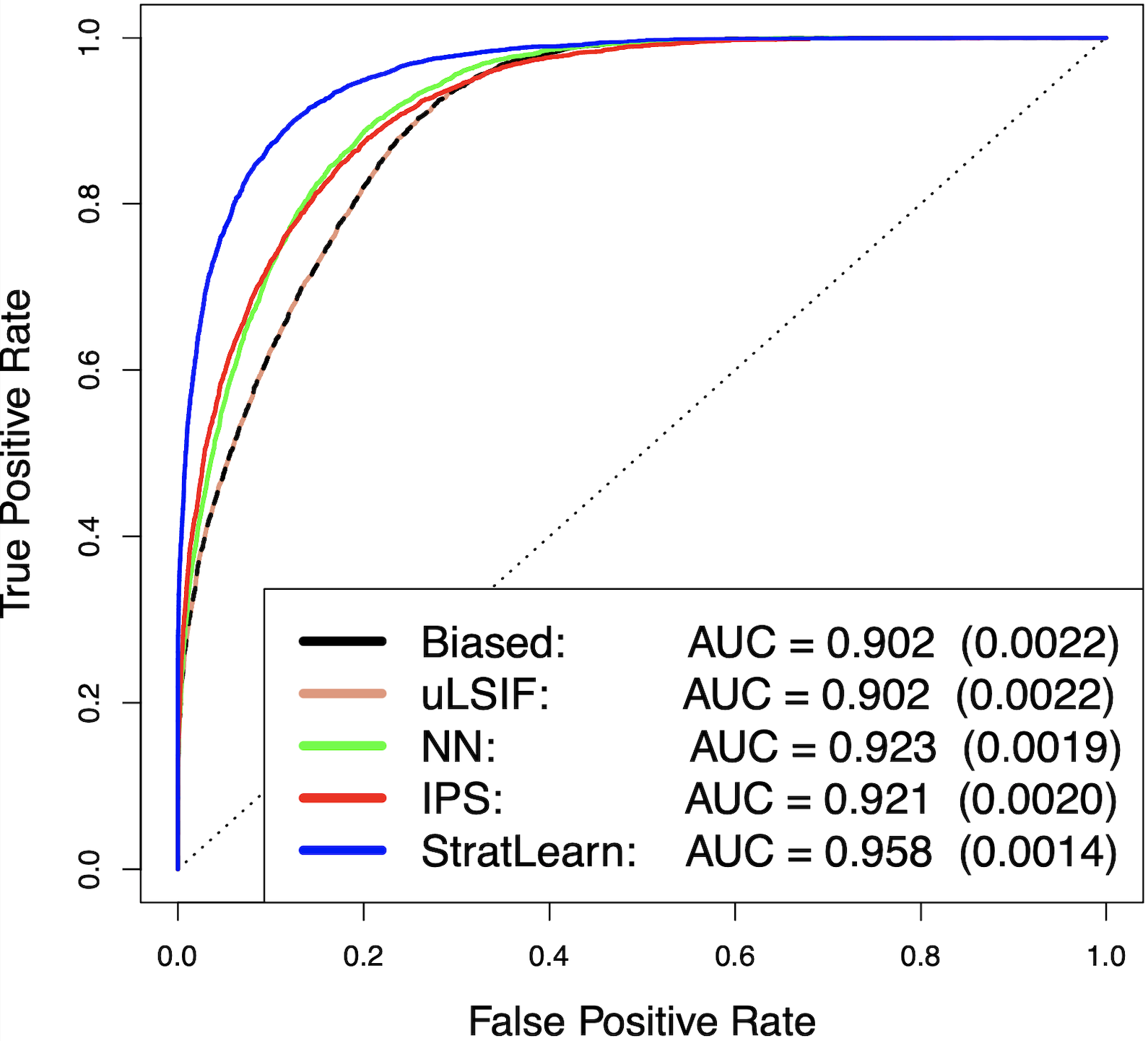}
  \caption{\baselineskip=15pt Comparison of ROC curves for SNIa classification using the updated SPCC data. Here, Biased and uLSIF are identical. Bootstrap AUC standard errors (from 400 bootstrap samples) are given in parentheses. \label{figure:AUC_updatedSPCC}}
\end{minipage}%
\hspace{0.3cm}
\begin{minipage}{.47\textwidth}
  \centering
  \vspace{-1.1cm}
  \includegraphics[width=0.95\linewidth]{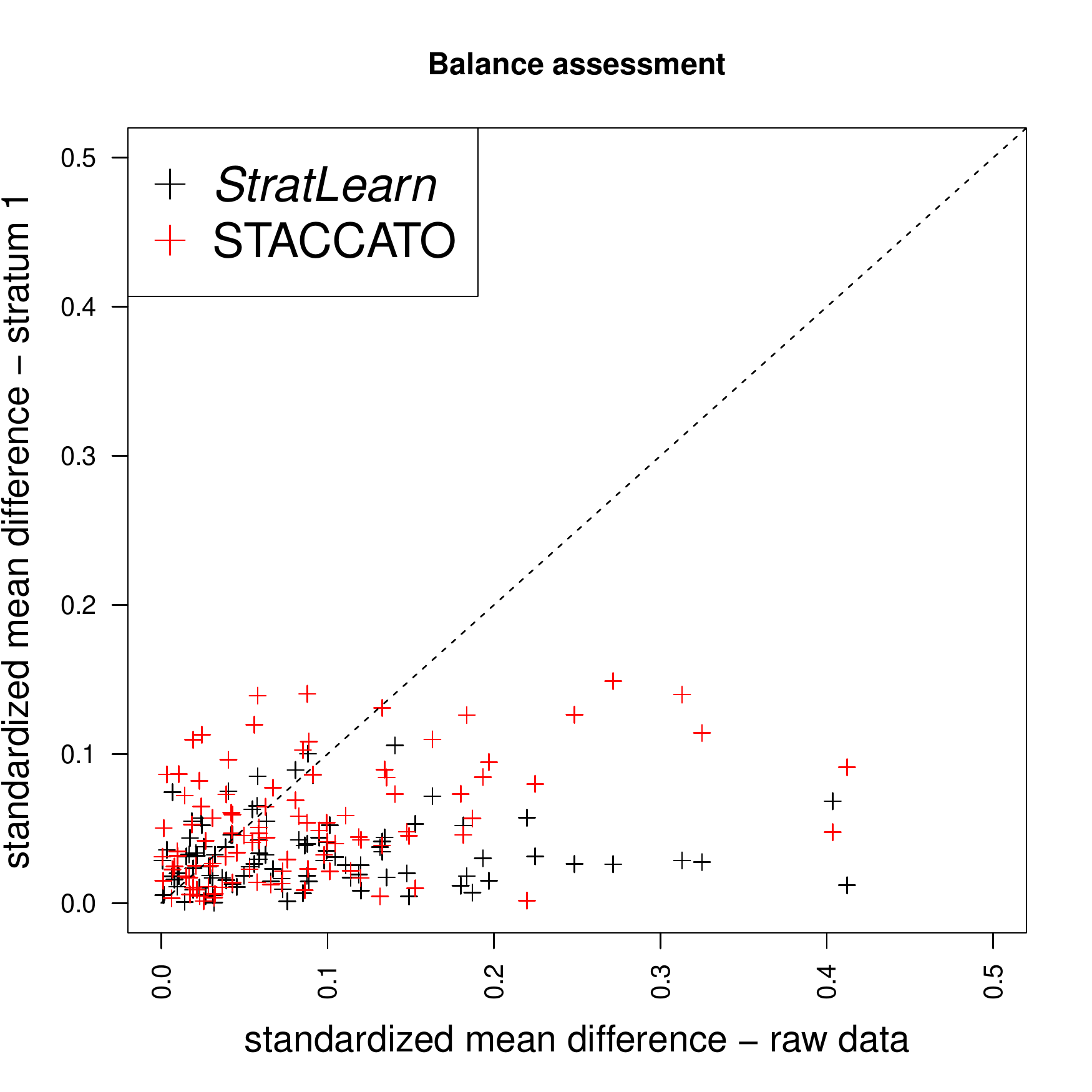}
		\captionof{figure}{
		\baselineskip=15pt Absolute standardized mean differences between source and target data of stratum 1 plotted against ``raw" data absolute standardized mean differences for 
		\textit{StratLearn} and STACCATO. 
		\label{figure:smd-strata1}}
\end{minipage}
\end{figure}

To evaluate the impact of covariate shift on classification, 
we first consider a `biased fit' by training 
a random forest classifier (as in 
\cite{revsbech2018staccato}) on the source covariates ignoring covariate shift, resulting in an AUC of 0.902 on the target data (black ROC curve in Fig~\ref{figure:AUC_updatedSPCC}). 
Next, we obtain a `gold standard' benchmark by randomly selecting 1102 objects from 
target data as a representative source set.
The same classification procedure with the unbiased `gold standard' training data (unavailable in practice) yields an AUC of 0.972 
on the remaining 
19,114 target objects.

Given the biased source 
data, \textit{StratLearn} is implemented as described in Section~\ref{section:methodology}, including all 102 covariates in the logistic propensity score estimation model. 
After stratification, a random forest classifier is trained and optimized on source strata $D_{S_1}$ and $D_{S_2}$ separately to predict samples in 
target strata $D_{T_1}$ and $D_{T_2}$. 
We use repeated 10-fold cross validation with a large hyperparameter grid to minimize the empirical risk of (\ref{form:strata_risk}) within each strata, employing  log-loss\footnote{ The log-loss (also referred to as cross-entropy loss) compares the output of a classification $f(x) \in [0,1]$ with the true output $y$ for an observation $(x,y)$ via $\mathcal{\ell}_{\rm logloss}(f(x),y) :=  -(y \log(f(x)) + (1 -y) \log(1 - f(x))).$ } as our loss-function; details appear in Section~\ref{section:supplement_upSPCC}.
Source strata $D_{S_j}$ for $j \in \{3,4,5\}$ have a small sample size, (13,7,4) respectively. Thus, source strata 
$D_{S_j}$ for $j \in \{3,4,5\}$ are merged with $D_{S_2}$ to train the random forest to predict $D_{T_j}$ for $j \in \{3,4,5\}$. 
With \textit{StratLearn}, we obtain an AUC of 0.958 on the target data (blue ROC curve in Fig~\ref{figure:AUC_updatedSPCC}), very near the optimal `gold standard' benchmark.

Fig~\ref{figure:AUC_updatedSPCC} compares \textit{StratLearn} to importance sampling methods designed to adjust for covariate shift. For importance sampling, the bootstrapped samples in the random forest fit were resampled with probabilities proportional to the estimated importance weights 
(see Section~\ref{section:supplement_comparison_models}). 
NN and IPS led to the best importance weighted classifier (AUC = 0.923, 0.921) -- an improvement over the biased fit, but substantially lower than \textit{StratLearn}. AUC standard errors (see Fig~\ref{figure:AUC_updatedSPCC}) are small relative to the large performance improvement of \textit{StratLearn}.
KLIEP failed to fit importance weights and is thus not included in the results. We also 
implemented IWCV 
using the same hyperparameter grid as for \textit{StratLearn}, and a combination of IWCV and importance sampling, which both led to lower AUC than the ones reported in Fig~\ref{figure:AUC_updatedSPCC} (see Section~\ref{section:supplement_upSPCC}).

Previous state-of-the-art methods report an AUC of 0.855 \citep{lochner2016photometric} using boosted decision trees, 0.939 \citep{pasquet2019pelican} using a  framework of an autoencoder and a convolutional neural network
and 0.94 \citep{revsbech2018staccato} using LC augmentation and target data leakage, all lower than {\em StratLearn}.

\paragraph{Balance Assessment on updated SPCC data:}
To illustrate the balancing property of propensity scores (see Section~\ref{section:balance_diagnostics}) and its effect on predictive target performance, we assess the covariate balance in the updated SPCC data within strata conditional on the estimated propensity scores, by means of two commonly used balance measures: absolute standardized mean differences (SMD) and the Kolmogorov-Smirnov test statistics (KS-stats) 
\citep{Austin2011,austin2015moving}.

Fig~\ref{figure:smd-strata1} provides a detailed covariate balance comparison, 
by plotting the ``raw" SMD against the \textit{StratLearn} SMD in stratum 1 (black) for each covariate. 
We remove two outliers (redshift and brightness) 
with very large ``raw" SMD (1.1 and 1.7), because including them in the Figure makes it more difficult to illustrate the balance of the bulk of the covariates; both are well balanced
in stratum~1 using \textit{StratLearn} (SMD equals 0.12 and 0.17).
Points below the diagonal line are better balanced in the stratum 
than in the ``raw" non-stratified data.
This is the case for the vast majority (71\%) of black points in Fig~\ref{figure:smd-strata1}, illustrating the balance improvement achieved with \textit{StratLearn}.

Fig~\ref{figure:smd-strata1} also plots (red) 
the SMD achieved by STACCATO \citep{revsbech2018staccato}, which uses two covariates (redshift and brightness, as opposed to the 102 used by \textit{StratLearn}) in the logistic regression to estimate propensity scores. 
While STACCATO improves the balance of the majority (69\%) of the covariates, most (66\%)  black (\textit{StratLearn}) SMD have smaller vertical values, indicating better balance than STACCATO (red).

On average across the 102 covariates, \textit{StratLearn} improves covariate balance compared to the ``raw" non-stratified data measured by SMD by ${\sim}70\%$ in stratum 1 and ${\sim}10\%$ in stratum 2 (KS-stats: ${\sim}70\%$ in stratum 1 and ${\sim}30\%$ in stratum 2)\footnote{Percentages are calculated by taking the ratio of the average SMD (average KS-stats) of all 102 covariates.}.
It further improves upon STACCATO by ${\sim}36\%$ in stratum 1 and ${\sim}46\%$ in stratum 2 using SMD (KS-stats: ${\sim}24\%$ in stratum 1 and ${\sim}36\%$ in stratum 2). The remaining strata contain too few source data to assess covariate balance. Details are provided in 
Table~\ref{table:smd_values_updated}.

The improved covariate balance (reduced covariate shift) directly translates into improved predictive performance. STACCATO (including data augmentation and target data leakage) yields a target AUC of 0.94, 
whereas with \textit{StratLearn} we obtain a target AUC of 0.958 (without data augmentation and no target data leakage) -- a substantial improvement resulting from the improved covariate balance by accounting for potentially confounding covariates. 
In general, we note that balance is particularly important for covariates that are strongly predictive for the outcome. Domain-specific expertise might be necessary to identify such covariates in the individual cases in practice.
In Section~\ref{section:supplement_UCI}, we demonstrate how covariate balance can be improved by adjusting the propensity score model.

\begin{table}[t!]
		\caption{\baselineskip=15pt Strata composition on the updated SPCC data (Section~\ref{section:Classification_SPCC}), applying STACCATO (left) and \textit{StratLearn} (right). The number of SNe, as well as the number and proportion of SNIa are presented in source and target stratum 1 and 2. For conciseness, we present the combined strata 3 to 5, containing too little source data for meaningful comparison of the SNIa proportions.	
	 \label{table:strata_proportions_updatedSPCC}}
\begin{tabular}{*{9}{c}} 
\hline
&& &STACCATO & &&& \textit{StratLearn}& \\
\cline{3-5} \cline{7-9}
\multirow{2}{*}{} & & Number & Number & Prop.& & Number & Number & Prop. \\
Stratum & Set & of SNe & of SNIa & of SNIa & & of SNe & of SNIa & of SNIa \\ 
\hline
\hline
\multirow{1}{*}{} 1 & Source &  924 & 414 & \textbf{0.45}&& 958 & 518 & \textbf{0.54}\\ 
& Target & 3340 & 1125 & \textbf{0.34} && 3306 & 1790 & \textbf{0.54} \\ 
\hline
\multirow{1}{*}{} 2 & Source & 153 & 125 & \textbf{0.82} && 120 & 28 & \textbf{0.23} \\ 
& Target &  4111 & 973 & \textbf{0.24}&& 4144 & 927 & \textbf{0.22} \\ 
\hline
\multirow{1}{*}{}3 to 5 & Source & 25 & 19  & 0.76 &&  24 & 12 & 0.5  \\ 
& Target &12,765   & 2431   &  0.19&&   12,766 & 1812 & 0.14 \\ 
\end{tabular}
\end{table}

Table~\ref{table:strata_proportions_updatedSPCC} presents the composition of the five \textit{StratLearn} strata. Recall that according to Remark~\ref{remark:potential_outcomes}, conditional on the propensity score the marginal distributions of source and target outcome are the same in expectation. 
Table~\ref{table:strata_proportions_updatedSPCC} shows that the proportion of SNIa in the source and target data (which in this case can be computed from knowledge of the true target labels in the simulation)  
align well for \textit{StratLearn} in the first two strata, indicating the expected reduction in covariate shift. The source sample sizes in strata 3-5 are quite small, rendering meaningful comparison of the SNIa proportions impossible. 
In Strata 1 and 2, however, \textit{StratLearn} achieves much better balance than either STACCATO or the raw (un-stratified) data (51\% SNIa in source, 23\% SNIa in target).

In Table~\ref{table:strata_predicted_proportions_updatedSPCC}, we demonstrate how predicted outcomes can be employed for balance diagnostics 
by assessing the predicted proportions of SNIa within strata obtained by STACCATO and by \textit{StratLearn}. 
We compute the predicted outcomes by classifying objects to be SNIa if the (random forest) predicted SNIa probabilities are above 0.5.
While STACCATO leads to a strong discrepancy between predicted SNIa proportions in the first two strata (indicating remaining confounding), \textit{StratLearn} leads to well-matched predicted SNIa proportions. 
We further quantify the discrepancy by performing a two-sided Fisher's exact test of independence, with the null hypothesis that there is no association of source/target set assignment and predicted SNIa proportion. 
Comparing different propensity score models, a higher p-value is an indicator for better balance in the predicted outcomes, and should thus be desirable.
\textit{StratLearn} leads to much higher p-values than STACCATO (failing to reject the null hypothesis for reasonable significance levels), which implies much weaker relation between source/target assignment and predicted outcomes. 

In this particular example, with \textit{StratLearn}, we fail to reject the null hypothesis for most significance levels. This may not always be the case (e.g., Section~\ref{section:supplement_originalSPCC}). 
We recall that the strategy of conditioning on propensity scores via stratification leads to subgroups with similar (not identical) propensity scores, and thus to similar (not identical) joint distributions within strata (this is the approximation in (\ref{form:stratified_joint_dist})). This in turn might lead to differences in the distributions of the covariates and the (predicted) outcomes, even if we could condition on the true propensity scores.
We thus employ the p-values of (predicted) outcomes as an additional tool to asses, and primarily to compare, propensity score models 
to detect and reduce confounding of highly predictive and thus most relevant covariates.

\begin{table}[t!]
		\caption{\baselineskip=15pt Outcome balance diagnostics via predicted labels on the updated SPCC data (Section~\ref{section:Classification_SPCC}), applying STACCATO (left) and \textit{StratLearn} (right). The number and proportion of predicted SNIa are presented in source and target stratum 1 and 2. P-values are computed via Fisher's exact test of independence between predicted SNIa target and source proportions within strata. 	
	 \label{table:strata_predicted_proportions_updatedSPCC}}
\begin{tabular}{*{9}{c}} 
\hline
&& \multicolumn{2}{c}{STACCATO (predicted) }  &&&  \multicolumn{2}{c}{\textit{StratLearn} (predicted) } &\\
 
\cline{3-5}
\cline{7-9}
\multirow{2}{*}{} & & Number &  Prop. & p-value & & Number &  Prop. & p-value  \\
Stratum & Set  & of SNIa & of SNIa &  & & of SNIa & of SNIa &  \\ 
\hline
\hline
\multirow{1}{*}{1} & Source &  414 & \textbf{0.45}&8.4e-11 && 518 &   \textbf{0.54} & 0.284\\ 
& Target & 1106 &  \textbf{0.33}& && 1853 & \textbf{0.56} & \\ 
\hline
\multirow{1}{*}{2} & Source & 125 & \textbf{0.82}& 2.8e-13 && 28 &  \textbf{0.23} & 0.749\\ 
& Target &  2166 &  \textbf{0.53}& && 1040 & \textbf{0.25} & \\ 
\hline
\end{tabular}
\end{table}


\subsection{Conditional Density Estimates -- Photo-$z$  Regression}\label{section:photoZ}

\paragraph{Objective:}

The wavelength of light from extragalactic objects is stretched because of the expansion of the universe -- a phenomenon called `redshift'. This fractional shift towards the red end of the spectrum 
is denoted by $z$. A precise measurement of redhsift allows cosmologists to estimate distances to astronomical sources, and its accurate quantification is essential for cosmological inference (e.g., redshift is a key component of the Big Bang theory).
Because of instrumental limitations, redshift can be precisely measured only for a small fraction of the $\sim 10^7$ galaxies observed to date (set to grow to $\sim 10^9$ within a decade). These source data 
are subject to covariate shift relative to the set of galaxies with unknown redshift (target). 
\cite{izbicki2017photo} employed importance weighting to adjust for covariate shift in $x$, a set of observed photometric magnitudes (a logarithmic measure of passband-filtered brightness), when estimating $z$. They obtain a non-parametric estimate of the full conditional density, $f(z|x)$, 
to quantify predictive uncertainty of redshift estimates. Proper quantification of predictive uncertainties is crucial to avoid systematic errors in the scientific downstream analysis \citep{izbicki2017photo,sheldon2012photometric}.
Using the same setup and conditional density 
estimation models (hist-NN, ker-NN, Series and Comb, detailed in \cite{izbicki2017photo}),\footnote{ \href{https://projecteuclid.org/journals/annals-of-applied-statistics/volume-11/issue-2/Photo-z-estimation--An-example-of-nonparametric-conditional-density/10.1214/16-AOAS1013.full?tab=ArticleLinkSupplemental}{For the computation of the conditional density estimators we used code by~\cite{izbicki2017photo} (link).} \label{footnote:photoU_conDens} } we show that \textit{StratLearn} leads to better overall predictive performance than importance weighting. 

Assuming that source and target data follow the same distribution, under the $L^2-$loss, conditional density estimators typically aim to minimize the {\it generalized} risk (generalized in that the underlying loss can be negative):
\begin{align} \label{formula:SDSS_source_loss_labelled}
    \hat{R}(\hat{f}) = & \frac{1}{n_S} \sum_{k=1}^{n_S} \int \hat{f}^2 (z|x_S^{(k)}) dz  -  2 \frac{1}{n_S} \sum_{k=1}^{n_S}  \hat{f} (z_S^{(k)}|x_S^{(k)}), 
\end{align} 
\cite{izbicki2017photo} propose to adjust for covariate shift by adapting (\ref{formula:SDSS_source_loss_labelled}), via optimizing weighted versions of the conditional density estimators \citep[Sections~5.1-5.3]{izbicki2017photo} with respect to an importance weighted {\it generalized} risk:
\begin{align} \label{formula:SDSS_source_loss}
    \hat{R}_S(\hat{f}) = & \frac{1}{n_T} \sum_{k=1}^{n_T} \int \hat{f}^2 (z|x_T^{(k)}) dz  -  2 \frac{1}{n_S} \sum_{k=1}^{n_S}  \hat{f} (z_S^{(k)}|x_S^{(k)}) \hat{w}(x_S^{(k)}), 
\end{align} 
where the weights, $\hat{w}(x_S) = p_T(x)/p_S(x)$, are estimated using the methods 
described in Section~\ref{section:comparison_models}. As their best performing model 
for $f(z|x)$, \cite{izbicki2017photo} propose an average of importance weighted ker-NN and Series, 
\begin{align} \label{formula:comb}
    \hat{f}^{\alpha}(z|x) = \sum_{k=1}^p \alpha_k \hat{f}_k(z|x),  \text{ with constraints } 
    \text{(i): } \alpha_i \geq 0,  \text{ and (ii): }  \sum_{k=1}^p \alpha_k = 1,
\end{align}
referred to as `Comb' (i.e., combination), where $p=2$ and $\alpha_i $ is optimized 
to minimize (\ref{formula:SDSS_source_loss}).

With \textit{StratLearn}, we optimize the unweighted conditional density estimators (hist-NN, ker-NN, Series) by minimizing (\ref{formula:SDSS_source_loss_labelled}) in each source stratum separately (accounting for covariate shift following Proposition~\ref{prop:cond_PS}). We also propose a \textit{StratLearn} version of Comb by optimizing (\ref{formula:comb}) on each source stratum separately (via the generalized risk in (\ref{formula:SDSS_source_loss}) with $w(x) \equiv 1 $), including ker-NN and Series (each optimized via \textit{StratLearn} beforehand).  
\textit{StratLearn} 
and the other methods are compared with a `Biased' 
(unweighted) method 
that simply optimizes (\ref{formula:SDSS_source_loss_labelled}). 
We abbreviate the combination of each method (\textit{StratLearn}, `Biased', and each of the weighting methods in Section~\ref{section:comparison_models}) with the models for $f(z|x)$ (hist-NN, ker-NN, Series and Comb) as $\text{Method}_{\text{Model}}$.

\paragraph{Data:}
We use the same data as \cite{izbicki2017photo}, consisting of 
467,710 galaxies from \cite{sheldon2012photometric}, each with spectroscopic redshift $z$ (measured with negligible error), and five photometric covariates $x$. As in \cite{izbicki2017photo}, we use the r-band magnitude and the four colors (differences of magnitude values in adjacent photometric bands) as our covariates.  We denote this spectroscopic source sample by $D_S$.
To simulate realistic covariate shift, we follow \cite{izbicki2017photo}: starting from $D_S$, we use rejection sampling to simulate a photometric, unrepresentative target sample $D_T$, with the prescription $p(s = 0 | x ) = f_{B(13,4)}(x_{(r)})/ \max_{x_{(r)}}  f_{B(13,4)}(x_{(r)})$, where $x_{(r)}$ is the $r$-band magnitude and $f_{B(13,4)}$ is a beta density with parameters (13,4). Additionally, we investigated adding $k \in \{10,50\}$ i.i.d. standard normal 
covariates as potential predictors to the 5 photometric covariates. This simulates the realistic case where additional potentially confounding covariates are present. 
For comparability, we follow \cite{izbicki2017photo} and use $|D_S^{\text{train}}| =2800$ galaxies randomly sampled from $D_S$ as training data, plus a validation set of $|D_S^{\text{val}}| = 1200$ galaxies. 
We assess the performance of each $\text{Method}_{\text{Model}}$ pair using a random subset of $D_T$, i.e., $ |D_T^{\text{test}}| = 6000$.

\begin{figure}[t!]
\centering
\begin{minipage}[H!]{.95\textwidth}
  \centering
    \includegraphics[width=\linewidth]{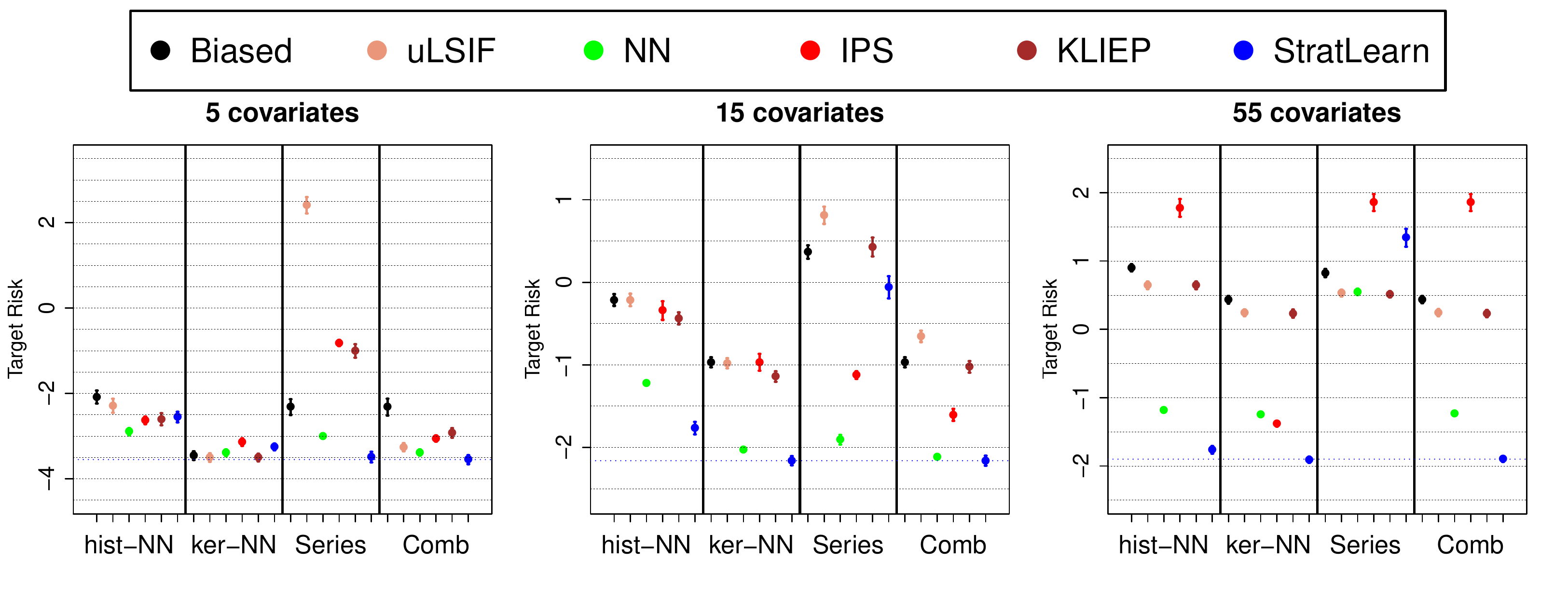} \caption{\baselineskip=15pt Target risk ($\hat{R}_T$) of the four 
    photo-$z$ estimation models under each 
    method (different colors), 
    using different sets of predictors. Bars give the mean $\pm$ 2 bootstrap standard errors (from $400$ bootstrap samples). 
    } 
    \label{figure:photoZ_results}
\end{minipage}
\end{figure}

\paragraph{Results:}  For evaluation of $\hat{f}$ under each 
$\text{Method}_{\text{Model}}$ pair, we use the (in our simulation) known target redshifts, $z_T$, to compute the target risk, $\hat{R}_T(\hat{f})$, via a non-weighted version of (\ref{formula:SDSS_source_loss}) with $x_S^{(k)}$ and $y_S^{(k)}$ replaced by $x_T^{(k)}$ and $y_T^{(k)}$, given by: 
\begin{equation}\label{formula:SDSS_target_loss}
    \hat{R}_T(\hat{f}) = \frac{1}{n_T} \sum_{k=1}^{n_T} \int \hat{f}^2 (z|x_T^{(k)}) dz  -  2 \frac{1}{n_T} \sum_{k=1}^{n_T}  \hat{f} (z_T^{(k)}|x_T^{(k)}).  
\end{equation} 
Fig~\ref{figure:photoZ_results} compares the resulting target risk $\hat{R}_T$ across models and covariate sets, showing that $\textit{StratLearn}_{\textit Comb}$ gives the best performance in all three covariate setups. 

For small covariate space dimension (Fig~\ref{figure:photoZ_results}, left panel), $\textit{StratLearn}_\textit{Comb}$ improves upon  $\textit{StratLearn}_{\textit{ker-NN}}$ and $\textit{StratLearn}_{\textit{Series}}$, optimizing the source risk in each stratum separately and combining their predictions. In the presence of potential additional confounding covariates (Fig~\ref{figure:photoZ_results}, middle and right panels), the performance of the Series estimator degrades strongly under most methods.
In these cases, $\textit{StratLearn}_{Comb}$ exploits the higher performance of $\textit{StratLearn}_{\textit{ker-NN}}$. 
In contrast, for the non-adjusted (Biased) and importance weighted methods (e.g. IPS), the combination of approaches (Comb) 
does not necessarily lead to improved performance (e.g.,
$\text{IPS}_{\text{Comb}}$ exhibits a higher target risk than $\text{IPS}_{\text{ker-NN}}$ on its own (Fig~\ref{figure:photoZ_results}, right panel)), 
indicating that the optimization in (\ref{formula:comb}) fails due to remaining covariate shift in the data. More precisely, the weighted empirical \textit{source} risk minimization ((\ref{formula:SDSS_source_loss}), as a form of (\ref{form:risk_shimodaira})) does not lead to \textit{target} risk minimization in these situations. 
In general, the improvement of \textit{StratLearn} relative to 
weighting
methods increases with the dimensionality of the covariate space, leading to a more robust regime.

\section{Discussion}
\label{section:discussion}


We provide a simple, though statistically principled and theoretically justified method for learning under covariate shift conditions.
We show that \textit{StratLearn} outperforms a range of state-of-the-art importance weighting methods on two contemporary research questions in cosmology (and on toy covariate shift examples, Section~\ref{section:supplement_UCI}), especially in the presence of a high-dimensional covariate space. 
The assumption of covariate shift is rather strong, requiring that there are no unmeasured confounding covariates -- something that cannot be guaranteed in general. In Section~\ref{section:supplement_UCI}, however, 
we demonstrate a certain robustness of our method against violation of this assumption. 
Further work is necessary to assess more fully the performance of \textit{StratLearn} when this assumption is only approximately fulfilled.
We emphasize that the covariate shift framework is best justified in the presence of a large number of covariates mitigating the risk of unmeasured confounders -- in which case it is critical to adopt a method that, like \textit{StratLearn}, can robustly handle many covariates. 
Our framework is entirely general
and versatile, as illustrated with examples of regression, conditional density estimation and classification.
Notably, our numerical demonstrations illustrate the advantage of using only a 
subset of the source data when formulating predictions for individual objects 
in the target, where the subset is chosen for its similarity to the target data in question (through stratification). This is a markedly different strategy to the widespread practice of including all possible available observations when fitting learning models.

The novelty of our approach is grounded in the transfer of the well-established causal inference propensity score framework \citepalias{rosenbaum1983central} 
to the domain adaptation/covariate shift setting, by demonstrating that a method established to obtain unbiased treatment effect estimates can be adapted 
to optimize the target risk of a supervised learner under covariate shift. 
In future work, this extension gives a chance to transfer 
hard won practical advice from causal inference (e.g., balance diagnostics, estimation of propensity scores, choice of included covariates \citep{rosenbaum1984reducing,Austin2007,Pirracchio2014}, etc.) to the covariate shift framework.
We will also explore the possibility of taking advantage of Proposition~\ref{prop:cond_PS} through a matching approach \citep{imbens2015causal}, 
which could prove more sensitive to the underlying propensity score distribution. 
We believe \textit{StratLearn} may become a powerful alternative to importance weighting, with a myriad of possible extensions and applications.

\section{Acknowledgements:}
David van Dyk acknowledges partial support from the UK Engineering and Physical Sciences Research Council [EP/W015080/1]; Roberto Trotta's work was partially supported by STFC in the UK [ST/P000762/1,ST/T000791/1]; and David Stenning acknowledges the support of the Natural Sciences and Engineering Research Council of Canada (NSERC) [RGPIN-2021-03985]. Finally, van Dyk, Stenning and Autenrieth acknowledge support from the Marie-Skodowska-Curie RISE [H2020-MSCA-RISE-2019-873089] Grant provided by the European Commission.
\textit{The authors report there are no competing interests to declare.}


\bibliography{./main.bbl}
\newpage


\setcounter{section}{0}
\renewcommand\thesection{S\arabic{section}}
\renewcommand{\theHsection}{S\arabic{figure}}

\setcounter{figure}{0}
\renewcommand\thefigure{S\arabic{figure}}
\renewcommand{\theHfigure}{S\arabic{figure}}

\setcounter{table}{0}
\renewcommand\thetable{S\arabic{table}}
\renewcommand{\theHtable}{S\arabic{figure}}

{\huge\bf Supplemental Materials}

\section{Additional Methodological Details for \textit{StratLearn}}\label{section:supplement_methodology}
In this section, we provide additional information for our novel \textit{StratLearn} methodology described in Section~\ref{section:methodology}, by formally deriving \Cref{prop:cond_PS}.
\bigskip
\\ \textbf{Verification of~\Cref{prop:cond_PS}:}
For \Cref{prop:cond_PS}, we start from 
[Theorem 1] in \cite{rosenbaum1983central}, which proves that the propensity score $e(x),$ with $0< e(x)<1,$ is a balancing score, which for our case means
\begin{align}\label{form:balancing_score}
x \ \Perp s | e(x),
\end{align} 
or equivalently $p_S(x | e(x)) = p_T(x | e(x))$. That is, conditional on the propensity score, the covariates in the source and target data have the same distribution. It follows that
\begin{align}
    p_S(x,y | e(x) ) :=& p(x,y | e(x), s = 1) \nonumber \\  \label{form:cond_PS1}
    =& p(y | x, e(x), s = 1) p(x| e(x), s = 1) \\ \label{form:cond_PS2}
    =& p(y | x, e(x), s = 0) p(x| e(x), s = 1) \\ \label{form:cond_PS3}
    =& p(y | x, e(x), s = 0) p(x| e(x), s = 0) \\  \label{form:cond_PS4}
    =& p(x,y | e(x), s = 0) \nonumber \\ 
    =:& p_T(x,y | e(x) ).  \nonumber
\end{align}
Step 
(\ref{form:cond_PS2}) follows from the covariate shift assumption that $p(y|x, s = 1) = p(y |x, s= 0)$, with $e(x)$ as a function of $x$ not changing the conditional distributions. Step 
(\ref{form:cond_PS3}) follows from the balancing property of the propensity score (\ref{form:balancing_score}). Thus, conditional on the propensity score, the source and target data have the same joint distribution. Equation~(\ref{form:exp_cond_ps}) follows directly.

\section{Bibliographic Note
}\label{section:supplement_bibnote}

In this section, we provide additional background literature to supplement examples provided in the main paper.

The purpose of domain adaptation methods is to obtain accurate target (test) predictions in situations where the source data domain is not an accurate representation of the target data domain.
\citesupp{quionero2009dataset} provide a comprehensive overview of such domain shift, or ``dataset shift", in machine learning. 
Domain adaptation cases arise in a myriad of applications, such as in medical imaging \citepsupp{stonnington2008interpreting,van2014transfer,kamnitsas2017unsupervised}, where mechanical configurations may vary between medical centers; natural language processing \citepsupp{jiang2007instance}, where annotated source data is often highly specialized and thus different from the target data; robotics and computer vision \citepsupp{patel2015visual,hoffman2016fcns,tai2017virtual,csurka2017comprehensive}, where simulated and observed data is combined to improve classification performance on the unlabeled target data;
and astronomy \citepsupp{gieseke2010detecting,vilalta2013machine,freeman2017unified,revsbech2018staccato,allam2018photometric,moller2020supernnova}, where brighter astronomical objects are more likely to be observed and therefore included in the source set. 

With increasing interest in the learning community, a variety of methods have been proposed, which (following \citesupp{kouw2019review}) can be mainly organized in three categories:
feature based methods, such as subspace mappings \citepsupp{fernando2013unsupervised,gong2013reshaping,kouw2016feature}, finding domain-invariant spaces \citepsupp{pan2010domain,gong2016domain} and optimal transport \citepsupp{courty2016optimal,courty2017joint};
inference based methods, including minimax estimators \citepsupp{liu2014robust,wen2014robust,chen2016robust} and self-learning \citepsupp{bruzzone2009domain,yoon2020vime}; and, thirdly, sample based approaches, with a focus on importance weighting \citepsupp{shimodaira2000improving,sugiyama2008direct,cortes2010learning,bickel2009discriminative}, mainly in the covariate shift framework.

The estimation of importance weights is central in the covariate shift framework, with a variety of different approaches.
\citesupp{yu2012analysis} and \citesupp{baktashmotlagh2014domain} estimate the densities separately, e.g. through kernel density estimators. 
Based on \cite{huang2007correcting}, \citesupp{wen2015correcting} propose Kernel-Mean-Matching in a reproducing kernel Hilbert space employing the frank-wolfe algorithm.
\citesupp{kanamori2012statistical} describe variations of unconstrained least-squares importance fitting (uLSIF) \citep{kanamori2009least}.
\citesupp{tsuboi2009direct} propose a variation of KLIEP \citep{sugiyama2008direct}, adopting a log-linear function to model the importance weights.
To alleviate the poor performance issue of weighting in high-dimensional covariate spaces, \citesupp{sugiyama2010dimensionality} and \citesupp{sugiyama2011direct} propose the incorporation of dimensionality reduction procedures when estimating density ratios. 
In theoretical work, \citesupp{ben2007analysis} and \citesupp{ben2010theory} derive generalization bounds, and conditions under which a classifier can learn from source data to perform well on target data. \citesupp{cortes2008sample} theoretically analyze the effect of errors in the weight estimation on the outcome hypothesis of the learning algorithm.
More recently, there have been efforts to transfer methods between the causal inference and domain adaptation framework \citepsupp{rojas2018invariant,magliacane2018domain,hassanpour2019counterfactual}.

In the causal inference framework, the introduction of propensity scores \citep{rosenbaum1983central} has been groundbreaking. 
There has been vast work on
generalization, best-practice, and assessment of propensity scores in causal inference \citepsupp[e.g.,][]{hirano2003efficient,imai2004causal,lunceford2004stratification,stuart2013prognostic,franklin2014metrics,austin2015moving,griffin2017chasing}.
While the general methodology has not yet been considered in the domain adaptation framework, it has been transferred to other areas, such as learning from positive and unlabeled data \citepsupp{bekker2019beyond}, unbiased learning-to-rank  \citepsupp{joachims2017unbiased}, or the evaluation of recommender systems \citepsupp{schnabel2016recommendations}, supplementing the examples provided in Section~\ref{sec:introduction}.
Much work has been done to improve propensity score estimators to obtain unbiased treatment effects, with various proposed methods  \citepsupp[e.g.,][]{Mccaffrey2004propensity,Setoguchi2008,Lee2010,westreich2011role,Mccaffrey2013,Pirracchio2014,imai2014covariate,Pirracchio2018,autenrieth2021stacked}. In the examples provided in the main paper, we show that \textit{StratLearn} with a simple logistic regression propensity score estimator leads to strong predictive target performance, outperforming various importance weighting methods. The consideration of additional propensity score estimators, as well as optimizing the set of selected covariates, might further improve \textit{StratLearn} in future work, e.g. by directly targeting covariate balance, similar to \citepsupp{griffin2017chasing,Parast2017,Pirracchio2018}.

We note that the additional literature described in this section is by far not complete, but rather serves as an additional overview of the domain adaptation/covariate shift framework, and allows further placement of our work within this framework. We refer to surveys provided in the literature for a more thorough discussion on the topic \citepsupp{pan2009survey,kouw2019review}. The summary of propensity score methods serves as an overview of the extensive work that can potentially be transferred from causal inference to the covariate shift framework, by transferring the general propensity score methodology as described in Section~\ref{section:methodology}. We refer to \cite{imbens2015causal} for a more comprehensive summary on propensity scores.

\section{Details of Comparison Models
}\label{section:supplement_comparison_models}

Here we detail our implementation of the several methods listed in Section~\ref{section:comparison_models} that we compare with \textit{StratLearn} in our numerical studies.

\paragraph{Importance Weighting Estimators:}
Following \citesupp{kanamori2009least} and \citesupp{bickel2009discriminative}, by
Bayes Theorem, 
\begin{equation} \label{form:weights_PS}
 w(x) = \frac{p_T(x)}{p_S(x)} = \frac{p(s = 1)}{p(s = 0)} \frac{p(s = 0 | x) }{p(s = 1 | x) } \approx \frac{n_S}{n_T} \left( \frac{1 }{p(s = 1 | x) } -1  \right).
\end{equation}

The importance-weighting-based estimators, KLIEP, uLSIF and NN (described in Section~\ref{section:comparison_models}) directly estimate $w(x)$. 
IPS obtains an estimate of $w(x)$ by plugging in the estimated propensity score into the right-hand side of (\ref{form:weights_PS}), which allows any probabilistic classifier to be used to obtain the weights, e.g., logistic regression \citepsupp{bickel2007dirichlet}. 
The estimated weights in (\ref{form:weights_PS}) can then be used for weighted empirical risk minimization (following Formula~(\ref{form:risk_shimodaira})); 
e.g., through (\ref{formula:SDSS_source_loss}) 
in Section~\ref{section:photoZ}, 
and through IWCV in  Section~\ref{section:Classification_SPCC}.

To implement importance sampling, such as in Section~\ref{section:Classification_SPCC}, we employ the framework proposed by
\cite{zadrozny2004learning}. Specifically, \cite{zadrozny2004learning} shows that for any distribution $\mathcal{D}$ with feature-label-selection space $\mathcal{X} \times  \mathcal{Y} \times  \mathcal{S}$ and $(x,y,s)$ examples drawn from  $\mathcal{D}$, for all classifiers $f$ and loss functions $\ell = \ell (f(x),y) $, if we assume that $ P(s = 1 | x,y) = P(s=1 |x) \neq 0$, then
\begin{align}\label{form:zadrozny1}
 \mathbb{E}_{\mathcal{D}} \left[\ell ( f(x),y)\right] =  \mathbb{E}_{\mathcal{\hat{D}}} \left[\ell ( f(x),y)| s = 1 \right], \\ 
 \text{with } \mathcal{\hat{D}}(x,y,s) :=  \frac{P(s= 1)}{P(s = 1| x)} \mathcal{D}(x,y,s).
\label{form:zadrozny2}
\end{align}
The target risk can thus be minimised by sampling from $\mathcal{\hat{D}}$, e.g., by importance sampling of source domain data.
Formula~(\ref{form:zadrozny2}) allows us to directly use the inverse of the estimated propensity scores. 
Moreover, rearranging the terms in the formulation of $w(x)$ in the right hand side of  (\ref{form:weights_PS}) yields
\begin{equation} \label{form:IPS}
    p(s = 0) w(x) + p(s=1)  = \frac{p(s = 1)}{p(s = 1|x)}.
\end{equation}
Then, substituting into (\ref{form:zadrozny2}),
\begin{align}
    \mathcal{\hat{D}}(x,y,s) :&=  \frac{P(s= 1)}{P(s = 1| x)}  \mathcal{D}(x,y,s) \label{form:PS_importance_sampling} \\
    & = \big(w(x) p(s = 0) + p(s=1) \big) \mathcal{D}(x,y,s).
    \label{form:w_importance_sampling}
\end{align}
Thus, following Formula~(\ref{form:zadrozny1}) \citep{zadrozny2004learning}, IPS can be used for importance sampling via (\ref{form:PS_importance_sampling}), and the direct-weight estimators KLIEP, uLSIF and NN can be used for importance sampling via (\ref{form:w_importance_sampling}).

\section{Details of Data and Software \label{section:supplement_data_software}}

The data required to reproduce the results for supernova classification, as presented in Section~\ref{section:Classification_SPCC} of the main paper, and Sections~\ref{section:supplement_upSPCC}~and~\ref{section:supplement_originalSPCC} of this Supplement, can be found here: 
\href{https://drive.google.com/drive/folders/1RIqhxDEP76SCvHeBhyf-3XtIODK5cytZ?usp=sharing}{SPCC data link}.
The data used for photometric redshift regression, as presented in Section~\ref{section:photoZ} was used by permission of the original authors of \cite{sheldon2012photometric}. The data used from the UCI repository \citepsupp{Dua:2019} is publicly available  (\href{https://archive.ics.uci.edu/ml/datasets/wine+quality}{UCI Wine quality data (link)}, 
\href{https://archive.ics.uci.edu/ml/datasets/parkinsons+telemonitoring}{Parkinsons Telemonitoring Data (link)}).

The \textit{StratLearn} software that we provide is written in R \citepsupp{R}. 
For computation of the experiments, a 32 CPU Core cluster 
was available to speed up the simulations.
However, computation time/resources was not a significant issue for the numerical results in this paper. As a benchmark, using a MacBook Pro with a 2.3 GHz 8-Core i9 processor, 
one scenario (e.g., medium covariate shift, high-dimensional covariate set) for conditional density estimation on SDSS data took less than 8 hours to run. All other experiments are less time-consuming. The supernova classification (Section~\ref{section:Classification_SPCC}) on preprocessed SPCC data (provided in the link above) 
took less than 2 hours on the same MacBook Pro.


\section{Additional Results for SNIa Identification \label{section:supplement_upSPCC}}

This Section provides additional results and details complementing Section~\ref{section:Classification_SPCC}, which describes the classification of supernovae (SNe) into type~1a (SNIa) and non-SNIa based on photometric light curve data, given a strongly biased source data set from \cite{kessler2010results}.

\paragraph{Random Forest Implementation and Hyperparameter Selection:} 
We use the ``ranger" random forest implementation in \texttt{caret}  \citepsupp{caret2008}
for SNe classification in this paper. 
Repeated 10-fold cross-validation (with five repetitions) is implemented with a large hyperparameter grid, containing four different numbers of covariates to possibly split at in each node (3,5,8,10); five values for minimum node size (3,5,7,10,20); and two different splitting rules (``gini" and ``extratrees").
To optimize the selected hyperparameter setting, we compute the average log-loss via cross-validation on source data. Using \textit{StratLearn}, cross-validation for hyperparameter selection (optimizing the average log-loss) is done on each source group separately (with groups defined in Section~\ref{section:Classification_SPCC}). We employ the log-loss to compute the empirical risk by evaluating the loss function $l(f(x),y)$ on each sample separately as described in Section~\ref{sec:preliminaries}.
The AUC (used to assess target predictive performance in this example) does not include the evaluation of a loss function that compares the true label $y$ and prediction $f(x)$ for each sample $x$ separately. However, to demonstrate robustness w.r.t. the choice of hyperparameter optimization, we also compute the AUC on source data (on each source group separately) via cross-validation, which leads to the same hyperparameter setting as using the log-loss.

\paragraph{Comparison of Importance Weighting Methods:} In addition to importance sampling (as presented in Figure~\ref{figure:AUC_updatedSPCC}), we investigate two further importance weighting approaches to obtain a comprehensive performance comparison of importance weighting and \textit{StratLearn}. More precisely, we implement IWCV, and a combination of IWCV and importance sampling. Table~\ref{table:importance_weighting_upSPCC} presents the resulting target AUC for all implemented weighting methods. In summary, none of the importance weighting methods are competitive with \textit{StratLearn}, which obtained a target AUC of 0.958. 
In fact, using IWCV alone leads to a decrease in  performance for NN (0.897) and IPS (0.897) 
relative to the `Biased' fit (0.902). 
In order to reweight the loss of each object separately when implementing IWCV, we optimize with respect to the log-loss instead of AUC. 
We do not expect this to be responsible for the decreased target performance. For example, using log-loss as the hyperparameter selection criterion for \textit{StratLearn} in each source group (instead of AUC) leads to the same target AUC of 0.958.
Importance sampling is the only weighting method for which we present the associated ROC curve in Figure~\ref{figure:AUC_updatedSPCC}, because it has the highest average AUC.

\begin{table}[bt]
\centering
\caption{AUC results on the updated SPCC (photometric) target data \protect\citep{kessler2010results}, using various importance weighting approaches to adjust for covariate shift. \label{table:importance_weighting_upSPCC} }
\begin{tabular}{lccc}
  \hline
 & IWCV & importance sampling & IWCV + importance sampling \\ 
  \hline
  \hline
uLSIF & 0.906 & 0.902 & 0.906 \\ 
  NN & 0.897 & 0.923 & 0.914 \\ 
  IPS & 0.897 & 0.921 & 0.924 \\ 
   \hline
\end{tabular}
\end{table}


\paragraph{Covariate Balance on Updated SPCC Data:}
Table~\ref{table:smd_values_updated} presents the covariate balance on the updated SPCC data before stratification (``raw"  data), and after stratification via STACCATO and \textit{StratLearn}, respectively.

\begin{table}[t!]
    \centering
    \caption{Average (SD) of SMD and KS-stats computed on the observed covariates from the updated SPCC data (diffusion map coordinates, redshift and brightness) on ``raw"  data, and in strata built by \textit{StratLearn} and STACCATO, respectively. Strata 3-5 are not displayed due to a shortage of source data. Smaller values indicate better balance.  
    \label{table:smd_values_updated} }
\begin{tabular}{l c c c c c}
\hline
    \multirow{2}{*}{ } & 
    \multirow{2}{*}{``raw"  data} &
   \multicolumn{2}{c}{STACCATO} &
     \multicolumn{2}{c}{\textit{StratLearn}} 
     \\
    \cmidrule(lr){3-4}
    \cmidrule(lr){5-6}
    & & Stratum 1 & Stratum 2 & Stratum 1 & Stratum 2  \\ 
    \hline
    \hline
    SMD       &  0.114 (0.202)  & 0.053 (0.039)  & 0.189 (0.195) &  0.034 (0.028) &   0.103 (0.161)\\
    KS-stats &  0.187 (0.084)   &  0.074 (0.05) & 0.206 (0.093)  &  0.056 (0.048) &  0.131 (0.08) \\
    \hline
\end{tabular}
\end{table}

On the non-stratified ``raw" data, we measure an average (SD) SMD of 0.114 (0.202) across the 102 covariates, with an average (SD) KS-stats of 0.187 (0.084). \textit{StratLearn} leads to increased balance within strata, strongly reducing the average (SD) SMD to 0.034 (0.028) in stratum 1; KS-stats: 0.056 (0.048); and to an average (SD) SMD of 0.103 (0.161); KS-stats:  0.131 (0.08); in stratum 2. (The remaining strata contain too few source data to assess covariate balance.) 

On average (SD) across all covariates, STACCATO led to an SMD of 0.053 (0.039); KS-stats: 0.074 (0.05), in stratum 1, and an average (SD) SMD of 0.189 (0.195), KS-stats: 0.206 (0.093), in stratum 2, much higher than \textit{StratLearn}.


\section{StratLearn applied to original SPCC Data 
}\label{section:supplement_originalSPCC}

This section provides additional numerical evidence for \textit{StratLearn}, performing SNIa classification on the original “Supernova photometric classification challenge” (SPCC) data set provided by \citep{kessler2010supernova}, an earlier version of the SPCC data described in Section~\ref{section:Classification_SPCC}.

\paragraph{Data:} 
The application of \textit{StratLearn} on the original SPCC data is mainly for comparison with previous literature that use this data set. The earlier version of the SPCC data features a total of 17,330 simulated SNe of type Ia (SNIa), Ib, Ic and II. The data set is divided into a source (training) set $D_S$ of 1217 spectroscopically confirmed SNe with known types, and (target set) $D_T$ of 16,113 SNe with unknown types and only photometric information. 
The simulation used to generate the original SPCC data suffered from a bug that made classification easier, thus leading to the updated SPCC data used in Section~\ref{section:Classification_SPCC}.
As in Section~\ref{section:Classification_SPCC}, we applied GP fitting and diffusion maps \citep{revsbech2018staccato,richards2012semi} to obtain a set of 102 predictive covariates; 100 covariates from the diffusion map, plus redshift (defined in Section~\ref{section:photoZ}) and a measure of the objects' brightness \citep{revsbech2018staccato}.

\begin{table}[tb]
\centering
\caption{Composition of the five strata on the original SPCC data \protect\citep{kessler2010supernova}. The number of SNe, as well as the number and proportion of SNIa are presented in each source and target stratum.\label{table:strata_proportions_originalSPCC}}
\begin{tabular}{ c c c c c } 
\hline
\multirow{2}{*}{} & & Number & Number & Prop. \\
Stratum & Set & of SNe & of SNIa & of SNIa \\ 
\hline
\hline
\multirow{1}{*}{1} & Source & 996 & 794 & 0.80 \\ 
& Target & 2470 & 1759 & 0.71 \\ 
\hline
\multirow{1}{*}{2} & Source & 210 & 56 & 0.27 \\ 
& Target & 3256 & 1010 & 0.31 \\ 
\hline
\multirow{1}{*}{3} & Source & 9 & 0 & 0 \\ 
& Target &  3457 & 385 & 0.11 \\ 
\hline
\multirow{1}{*}{4} & Source &  2 & 1 & 0.50 \\ 
& Target & 3464 & 258 & 0.07 \\ 
\hline
\multirow{1}{*}{5} & Source & 0 & 0 & NA  \\ 
& Target &  3466 & 180 & 0.05 \\ 
\hline
\end{tabular}
\end{table}

\paragraph{Results:} Table~\ref{table:strata_proportions_originalSPCC} presents the composition of the five strata obtained by conditioning on the estimated propensity scores, exhibiting a similar pattern as the strata composition on the updated SPCC data (Table~\ref{table:strata_proportions_updatedSPCC}), though with a higher proportion of SNIa in the first stratum and even less source data in strata $3-5$. 
We thus follow the same strategy as described in Section~\ref{section:Classification_SPCC}.
After stratification, a random forest classifier is trained and optimized on source strata $D_{S_1}$ and $D_{S_2}$ separately to classify SNe in 
target strata $D_{T_1}$ and $D_{T_2}$, respectively. Source strata $D_{S_j}$ for $j \in \{3,4,5\}$ are merged with $D_{S_2}$ to train the random forest to predict $D_{T_j}$ for $j \in \{3,4,5\}$. Repeated 10-fold cross validation (with the hyperparameter grid described in Section~\ref{section:supplement_upSPCC}) is performed to minimize the empirical risk of (\ref{form:strata_risk}) within each group. 

\begin{figure}[h!]
\centering
\begin{minipage}{.43\textwidth}
  \centering
    \includegraphics[width=\linewidth]{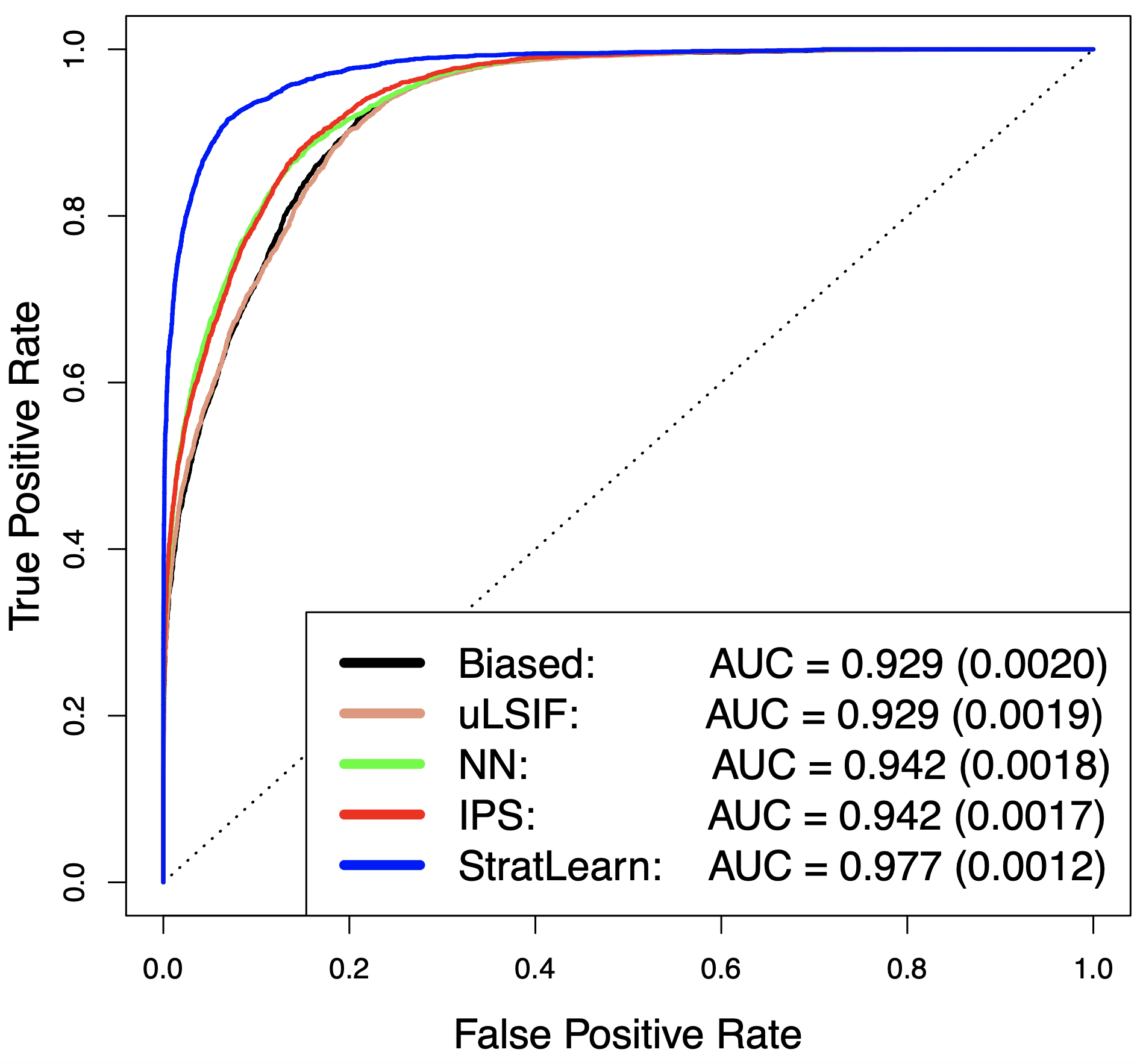}
\end{minipage}
 \caption{Comparison of ROC curves for SNIa classification using the original SPCC data \citep{kessler2010supernova}. Bootstrap AUC standard errors (from 400 bootstrap samples) are presented in parenthesis. \label{figure:AUC_originalSPCC}  }
\end{figure}

With \textit{StratLearn} we obtain an AUC of 0.977 on the target data of the original SPCC data (blue ROC curve in Figure~\ref{figure:AUC_originalSPCC}), very near the optimal `gold standard' benchmark, which is 0.981 
on this data set (obtained on a set with 1217 objects randomly selected from the target data as a representative source set, 
which would not exhibit covariate shift and would not be available in practice). The AUC obtained with \textit{StratLearn} is also larger than the previous best published AUC for this data, AUC = 0.961~\citep{revsbech2018staccato} with STACCATO, which employs a data augmentation strategy and target data leakage.
Note that the predictive target performance on the original SPCC data \citep{kessler2010supernova} is generally higher than on the updated version \citep{kessler2010results}, due to the aforementioned bug that made classification easier.

Most of the importance weighting methods (Table~\ref{table:importance_weighting_originalSPCC}) lead to slight improvements over the (non-adjusted) `Biased' model fit. The best performing weighting approaches (using estimated NN or IPS weights for importance sampling) obtain a target AUC of 0.942, substantially lower than \textit{StratLearn} (AUC: 0.977). Importance sampling is the only weighting method for which we present the associated ROC curve in Figure~\ref{figure:AUC_originalSPCC}, because it has the highest average AUC.


\begin{table}[t!]
\centering
\caption{AUC results on the original SPCC (photometric) target data \protect\citep{kessler2010supernova}, using various importance weighting approaches to adjust for covariate shift. \label{table:importance_weighting_originalSPCC} }
\begin{tabular}{lccc}
  \hline
 & IWCV & importance sampling & IWCV + importance sampling \\ 
  \hline
  \hline
uLSIF & 0.934 & 0.929 & 0.936 \\ 
  NN & 0.934 & 0.942 & 0.931 \\ 
  IPS & 0.934 & 0.942 & 0.931 \\ 
   \hline
\end{tabular}
\end{table}


\subsection{Covariate Balance Check on Original SPCC Data}
 
In this section, we illustrate the balancing property of propensity scores by assessing the covariate balance in the original SPCC data within strata conditional on the estimated propensity scores, such as described in the last paragraph of Section~\ref{section:methodology}. In particular, we show that such balance assessment could be helpful to assess the suitability of choice of covariates and/or the propensity score model.  

As mentioned in Section~\ref{section:Classification_SPCC}, the method proposed by \cite{revsbech2018staccato} (STACCATO) can be viewed as a prototype of \textit{StratLearn}, as it augments the spectroscopic source data separately in strata based on the estimated propensity scores. STACCATO employs two covariates (``redshift" and measure of brightness) as main effects in a logistic regression model to estimate the propensity scores. In contrast, \textit{StratLearn} includes all 100 diffusion map covariates (Section~\ref{section:Classification_SPCC}), in addition to ``redshift" and ``brightness", as main effects in a logistic regression model to estimate the propensity scores. 
We assess the balance achieved through stratification by means of two commonly used balance measures: absolute standardized mean differences and the Kolmogorov-Smirnov test statistics \citepsupp{Austin2011,Mccaffrey2013}, computed for each covariate within strata. Table~\ref{table:smd_values} shows that, on average across observed covariates, \textit{StratLearn} leads to substantially reduced absolute standardized mean differences and Kolmogorov-Smirnov test statistics in strata one and two, compared to STACCATO.
We only report balance measures in strata one and two because of a shortage of source data in the remaining strata. 
Thus, building the strata by conditioning on the \textit{StratLearn} estimated propensity scores leads to improved covariate balance (i.e., reduced covariate shift) within strata, compared to using the STACCATO estimated propensity scores. Notably, comparing the predictive performance, STACCATO (without augmentation) yields a target AUC of 0.942, whereas with \textit{StratLearn} we obtain an AUC of 0.977 on the target data -- a substantial improvement resulting from the improved covariate balance by accounting for potentially confounding covariates.

\begin{table}[t!]
    \centering
    \caption{Average (SD) of SMD and KS-stats computed on the observed covariates from the original SPCC data (diffusion map coordinates, redshift and brightness) on ``raw"  data, and in strata built by \textit{StratLearn} and STACCATO, respectively. Strata 3-5 are not displayed due to a shortage of source data. Smaller values indicate better balance.  
    \label{table:smd_values} }
\begin{tabular}{l c c c c c}
\hline
    \multirow{2}{*}{ } & 
    \multirow{2}{*}{``raw"  data} &
    \multicolumn{2}{c}{\textit{StratLearn}} &
    \multicolumn{2}{c}{STACCATO} 
     \\
    \cmidrule(lr){3-4}
    \cmidrule(lr){5-6}
    & & Stratum 1 & Stratum 2 & Stratum 1 & Stratum 2  \\ 
    \hline
    \hline
    SMD       & 0.126 (0.184)  & 0.031 (0.026) & 0.066 (0.121) & 0.057 (0.055) & 0.132 (0.127)\\
    KS-stats &  0.186 (0.073)  &   0.053 (0.019) & 0.095 (0.053)  & 0.084 (0.027) & 0.163 (0.066) \\
    \hline
\end{tabular}
\end{table}

\begin{figure}
\centering
\begin{minipage}{.5\textwidth}
  \centering
\includegraphics[width=0.9\linewidth]{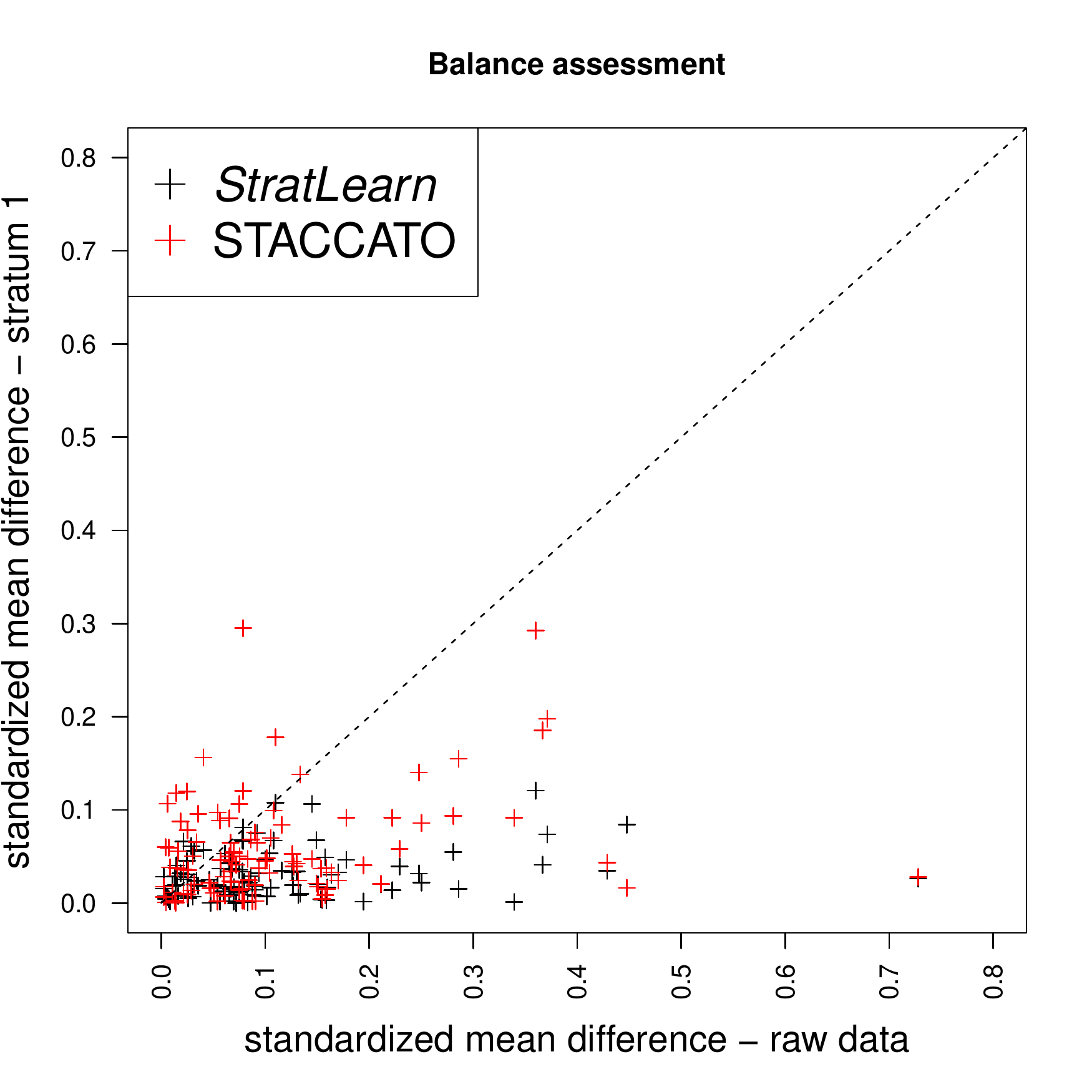}
\end{minipage}
  \caption{Absolute standardized mean differences between source and target data of stratum 1 plotted against raw data absolute standardized mean differences for both propensity score (PS) estimation approaches (\textit{StratLearn} and STACCATO). For visual clarity we left out one covariate which would appear at the coordinates (1.5,0.09). \label{figure:smd-strata1_original}}
\end{figure}

For illustration purposes, in Figure~\ref{figure:smd-strata1_original} we display a more detailed balance check, comparing the absolute standardized mean differences computed for each covariate in stratum 1 (obtained through \textit{StratLearn} (black) and STACCATO (red)) with 
the absolute standardized mean differences of the covariates without stratification. 
In general, points below the diagonal line indicate better balance in the stratum compared to the balance on the `raw' non-stratified data. 
Even though \cite{revsbech2018staccato} (STACCATO) just include redshift and brightness in the propensity score estimation, both approaches balance the majority of the covariates. However, the black (\textit{StratLearn}) standardized mean differences have visibly smaller values in the y-direction, indicating better balance compared to the approach in \cite{revsbech2018staccato} (red). Note that for clarity of Figure~\ref{figure:smd-strata1_original}, we removed one outlier (brightness) with a raw standardized mean difference of 1.5, which was very well balanced with a standardized mean difference of around 0.09 in stratum 1 by both approaches. 
Assessing the balance of individual covariates as illustrated in Figure~\ref{figure:smd-strata1_original} might be particularly useful to evaluate (and improve) the resulting balance of covariates important for outcome prediction. Some domain-specific expertise might be necessary to identify such covariates.
In Section~\ref{section:supplement_UCI}, we show how covariate balance assessment can be employed to evaluate the propensity score model choice.

Table~\ref{table:strata_predicted_proportions_originalSPCC} presents balance comparisons between STACCATO and \textit{StratLearn} via predicted outcome labels (as described in Section~\ref{section:balance_diagnostics}). \textit{StratLearn} leads to much better balance between predicted outcome proportions within strata one and two, and thus to much higher p-values than STACCATO, implying a much weaker relation between source/target assignment and predicted outcomes. 
The comparison of predicted outcome proportions and p-values (between \textit{StratLearn} and STACCATO) provide a straightforward indication to employ the \textit{StratLearn} propensity score model for stratification, which in turn leads to much improved outcome model predictive performance of SNIa classification.

\begin{table}[t!]
		\caption{\baselineskip=15pt Outcome balance diagnostics via predicted labels on the original SPCC data (Section~\ref{section:supplement_originalSPCC}), applying STACCATO (left) and \textit{StratLearn} (right). The number and proportion of predicted SNIa are presented in source and target stratum 1 and 2. P-values are computed via Fisher's exact test of independence between predicted SNIa target and source proportions within strata. 	
	 \label{table:strata_predicted_proportions_originalSPCC}}
\begin{tabular}{*{9}{c}} 
\hline
&& \multicolumn{2}{c}{STACCATO (predicted) }  &&&  \multicolumn{2}{c}{\textit{StratLearn} (predicted) } &\\
 
\cline{3-5}
\cline{7-9}
\multirow{2}{*}{} & & Number &  Prop. & p-value & & Number &  Prop. & p-value  \\
Stratum & Set  & of SNIa & of SNIa &  & & of SNIa & of SNIa &  \\ 
\hline
\hline
\multirow{1}{*}{1} & Source &  652 & \textbf{0.69}&5.0e-21 && 794 &   \textbf{0.80} & 0.013\\ 
& Target & 1290 &  \textbf{0.51}& && 1872 & \textbf{0.76} & \\ 
\hline
\multirow{1}{*}{2} & Source & 181 & \textbf{0.74}& 6.9e-23 && 56 &  \textbf{0.27} & 0.016\\ 
& Target &  1336 &  \textbf{0.41}& && 1132 & \textbf{0.35} & \\ 
\hline
\end{tabular}
\end{table}


\section{\textit{StratLearn} applied to Data from UCI Repository
}\label{section:supplement_UCI}

In this section, we demonstrate the performance of \textit{StratLearn} under violation of the covariate shift assumption.
We apply \textit{StratLearn} on two datasets from the publicly available UCI repository \citepsupp{Dua:2019}: the ``wine quality data" \citepsupp{cortez2009modeling} 
and the ``Parkinson Telemonitoring data" \citepsupp{little2008suitability}. 
These datasets have previously been used to explore regression problems in the presence of covariate shift \citepsupp{chen2016robust}.  Links to these data sets can be found in Section~\ref{section:supplement_data_software}.

\subsection{Data Description}

\paragraph{Wine Quality Data:} The data set contains 6497 samples of which 4898 are of type ``white wine", used as source data $D_S$, and 1599 samples are of type ``red wine", used as the target data $D_T$. The output is a wine ``quality score" (between 0 and 10), which we aim to predict for the target data (red wine), given the scores of the source data (white wine). There are 11  physicochemical, predictive covariates. We refer to \citesupp{cortez2009modeling} for detailed information.

\paragraph{Parkinson Data:}
The data comprises 5875 instances and 26 attributes. Similar to \citesupp{chen2016robust}, we take 17 of the attributes to be predictive covariates, with the aim of predicting the UPDRS Parkinson's disease symptom score (output). The source data $D_S$ contains all instances with age below 60, leading to a source subset of size $|D_S| = 1877.$ Instances with age above (inclusively) 60 and below 70 are used as target samples $D_T$, with $|D_T| = 2127.$ As in \citesupp{chen2016robust}, instances with age above 70 are not considered. We refer to \citesupp{little2008suitability} for further information on the data.

\subsection{Implementation and Results on UCI Data}
Following \citesupp{chen2016robust}, our objective is to predict the wine ``quality score" and the Parkinson's ``UPDRS'' score using linear regression models, specifically ordinary least squares (OLS). We assume
that the source and target splitting covariates (wine type and age) are not observed for model fitting. Note that,  because these are confounding covariates, this violates the covariate shift assumption. All other available covariates are used to estimate the propensity scores and the importance weights, and as predictor variables for the outcome models. To improve covariate balance on the wine quality data, we use gradient boosting machines to estimate the propensity scores, commonly used in the propensity score causal inference literature \citepsupp{Mccaffrey2004propensity,Mccaffrey2013,autenrieth2021stacked}. 
Covariate balance is assessed as described in the last paragraph of Section~\ref{section:methodology}. On the wine quality data, we assess the covariate balance in stratum 4, which is the only stratum with both source and target samples (Table~\ref{table:UCI_strata_composition}).
Using logistic regression, the average (SD) absolute standardized mean difference in the observed covariate set is 0.667 (0.436), and the average (SD) of the Kolmogorov-Smirnov test statistics is 0.388 (0.154). Estimating the propensity scores using gradient boosting machines increases balance in stratum 4, by reducing the average (SD) absolute standardized mean difference to 0.561 (0.292), and the average (SD) Kolmogorov-Smirnov test statistics to 0.288 (0.109). 
On the Parkinson data set, logistic regression is used to estimate the propensity scores because it leads to better balance in the strata than using gradient boosting machines.

Table~\ref{table:UCI_strata_composition} presents the composition of the \textit{StratLearn} strata for the Wine and Parkinson data, both conditioned on the estimated propensity scores. For the wine data set, all target samples are in strata 4 and 5, and only stratum 4 contains both source and target samples. Following our \textit{StratLearn} strategy, we fit OLS on $D_{S_4}$ to predict $D_{T_4}$ and $D_{T_5}.$ In the Parkinson data set there is enough source data in each strata to fit OLS separately on each source stratum $D_{S_j}$ to predict the data in the respective target stratum $D_{T_j}$, $ j \in \{1,\dots,5\}$.

Table~\ref{table:UCI_results} presents the MSE of the target predictions, comparing \textit{StratLearn} with (non-adjusted) OLS (`Biased') and WLS, using the methods described in Section~\ref{section:comparison_models} to estimate the weights. Applying \textit{StratLearn} results in a target MSE of 0.715 for the Wine data and 114.97 for the Parkinson data, both improving upon the `Biased' OLS fit which yields a target MSE of 1.024 (Wine) and 130.88 (Parkinson). \textit{StratLearn} further provides better results than WLS:uLSIF, WLS:KLIEP and WLS:NN on both data sets; only WLS:IPS leads to slightly better results than \textit{StratLearn}.

We also compare \textit{StratLearn} with the method that \citesupp{chen2016robust} develops and illustrates on these datasets. Specifically, \citesupp{chen2016robust} proposes a
``robust bias-aware regression'' based on the Kullback-Leibler divergence. This proposal is tailored specifically to regression and to loss functions that account for uncertainty in the prediction (e.g., empirical logloss), whereas \textit{StratLearn} is entirely general but can be successfully applied to this task. Following \citesupp{chen2016robust}, we compute the target empirical logloss as a performance measure. 
On the Wine data, \textit{StratLearn} yields an empirical logloss of 1.271, which exactly matches the reported performance of the method proposed in \citesupp{chen2016robust}. On the Parkinson data we could not replicate the exact data split as reported in \citesupp{chen2016robust}, and thus a meaningful comparison of empirical logloss to the values reported by \citesupp{chen2016robust} is not possible.

In summary, in these illustrative regression examples \textit{StratLearn} demonstrates certain robustness against violation of the covariate shift assumption, improving upon the (non-adjusted) `Biased' fit and performing comparably with the best importance weighting approach.

\begin{table}[t!]
\centering
\caption{Composition of the five \textit{StratLearn} strata for the UCI wine and UCI parkinson data. The number of samples/subjects in source and target stratum, as well as the mean outcome (``quality score" and ``UPDRS score" ) are presented. 
\label{table:UCI_strata_composition}}
\begin{tabular}{ l l r r  } 
\hline
\multirow{2}{*}{} &   &  UCI Wine data    & UCI Parkinson data  \\
Stratum & Set & \# samples (Mean ``quality") & \# subjects (Mean ``UPDRS") \\ 
\hline
\hline
\multirow{1}{*}{1} 
& Source & 1299 (5.98) & 627 (22.00) \\ 
& Target & 0 (0.00)  & 174 (29.15) \\ 
\hline
\multirow{1}{*}{2} 
& Source & 1300 (5.92) & 486 (26.36) \\ 
& Target & 0 (0.00)  & 315 (25.21) \\ 
\hline
\multirow{1}{*}{3} 
& Source  & 1300 (5.93) & 314 (25.53) \\ 
& Target & 0 (0.00) & 487 (24.86) \\ 
\hline
\multirow{1}{*}{4} 
& Source & 999 (5.63) & 269 (27.38) \\ 
& Target & 301 (5.49) & 532 (28.15) \\ 
\hline
\multirow{1}{*}{5} 
& Source & 0 (0.00) & 181 (27.06) \\ 
& Target & 1298 (5.67) & 619 (30.00) \\ 
\hline
\hline
\multirow{1}{*}{All} 
& Source & 4898 (5.88) & 1877 (24.98) \\ 
& Target & 1599 (5.64) & 2127 (27.58) \\ 
\end{tabular}  
\end{table}

\begin{table}[t!]
\centering
\caption{MSE of target predictions on UCI Wine and Parkinson data, based on ordinary least squares regression (OLS), various importance weighted least squares regression methods (WLS), and our proposed \textit{StratLearn} method. \label{table:UCI_results} }
\begin{tabular}{l r r}
  \hline
Method $\backslash$ Data &    UCI wine data   &  
  UCI Parkinson data \\
  \hline
  \hline
  OLS (Biased) & 1.024  & 130.88  \\ 
  WLS:uLSIF & 2.363 & 120.81  \\ 
  WLS:KLIEP & 3.968  & 116.72  \\ 
  WLS:NN & 2.377  & 117.47  \\ 
  WLS:IPS & 0.660 & 112.80  \\ 
  StratLearn & 0.715  & 114.97  \\ 
   \hline
\end{tabular}
\end{table}


\section{Additional Results for Redshift Regression 
\label{section:supplement_photoZ}}

In the context of the photometric redshift regression example in Section~\ref{section:photoZ}, we present additional numerical results that demonstrate the robustness of \textit{StratLearn} to varying degrees of covariate shift. Specifically, we reproduce the results of Section~\ref{section:photoZ}, but under two additional covariate shift scenarios. 

\paragraph{Additional Covariate Shift Setups:} 
We use the same data setup as in Section~\ref{section:photoZ}, but change the rejection sampling used to simulate an 
unrepresentative target $D_T$ from $D_S$, to 

\begin{itemize}
\item Weak covariate shift: \begin{equation}
    p(s = 0 | x ) = f_{B(9,4)}(x_{(r)})/ \max_{x_{(r)}}  f_{B(9,4)}(x_{(r)}) \label{form:SDSS_lowCS}
\end{equation} 
\item Strong covariate shift:  \begin{equation} p(s = 0 | x ) = f_{B(18,4)}(x_{(r)})/ \max_{x_{(r)}}  f_{B(18,4)}(x_{(r)}), \label{form:SDSS_strongCS}
\end{equation} 
\end{itemize}
where $x_{(r)}$ is the $r$-band magnitude and $f_{B(9,4)}$ and  $f_{B(18,4)}$ are beta densities with parameters (9,4) and (18,4), respectively. Except for the adjusted degree of covariate shift, simulations are performed as described in Section~\ref{section:photoZ}.

\begin{figure}[ht!]
\centering
\begin{minipage}[h]{.95\textwidth}
  \centering
    \includegraphics[width=\linewidth]{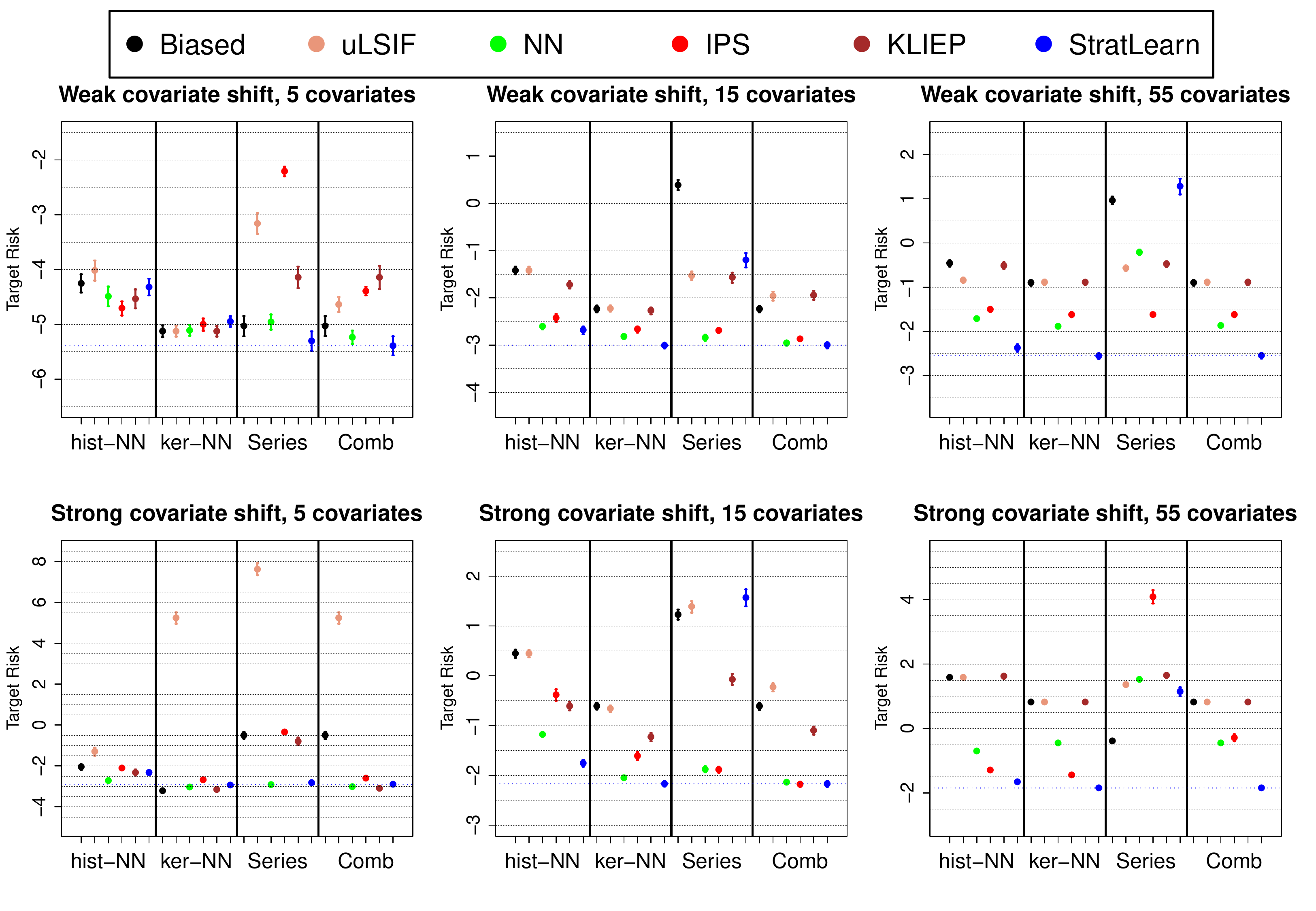} \caption{Target risk ($\hat{R}_T$) of the four photo-$z$ estimation models under each method (different colors), using different sets of predictors. Bars give the mean $\pm$ 2 bootstrap standard errors (from $400$ bootstrap samples).
    Top row: Weak covariate shift (following (\ref{form:SDSS_lowCS})); Bottom row: Strong covariate shift (following (\ref{form:SDSS_strongCS})). } \label{figure:append_photoZ_results}
\end{minipage}
\end{figure}

\begin{table}[t!]
\centering
\caption{Composition of the five \textit{StratLearn} strata for the medium covariate shift scenario (as described in Section~\ref{section:photoZ}) on SDSS data, using estimated propensity scores with different sets of predictors. The number of galaxies in source and target stratum, as well as the mean outcome (redshift $z$) are presented.}\label{table:sdss_strata_comp_medium_all}
\begin{tabular}{ l l r r r  } 
\hline
\multirow{2}{*}{} &   &  5 covariates    & 15 covariates & 55 covariates \\
Stratum & Set & \#galaxies (Mean $z$) & \#galaxies (Mean $z$) &\#galaxies (Mean $z$) \\ 
\hline
\hline
\multirow{1}{*}{1} 
& Source & 1631 (0.09) & 1583 (0.09) & 1620 (0.09) \\ 
& Target & 7 (0.07) & 9 (0.07) & 7 (0.07) \\  
\hline
\multirow{1}{*}{2} 
& Source & 1500 (0.13) & 1515 (0.13) & 1546 (0.13) \\  
& Target & 112 (0.11) & 113 (0.13) & 98 (0.11) \\ 
\hline
\multirow{1}{*}{3} 
& Source & 618 (0.29) & 641 (0.29) & 594 (0.30) \\ 
& Target & 1481 (0.33) & 1499 (0.33) & 1480 (0.33) \\ 
\hline
\multirow{1}{*}{4} 
& Source & 116 (0.42) & 114 (0.40) & 108 (0.40) \\ 
& Target & 2196 (0.39) & 2215 (0.38) & 2258 (0.38) \\ 
\hline
\multirow{1}{*}{5} 
& Source & 135 (0.48) & 147 (0.46) & 132 (0.47) \\ 
& Target & 2204 (0.48) & 2164 (0.48) & 2157 (0.48) \\ 
\hline
\hline
\multirow{1}{*}{All} 
& Source & 4000 (0.16) & 4000 (0.16) & 4000 (0.16) \\ 
& Target & 6000 (0.40) & 6000 (0.40) & 6000 (0.40) \\ 
\end{tabular}
\end{table}


\begin{table}[t!]
\centering
\caption{Composition of the five \textit{StratLearn} strata for the weak covariate shift (\ref{form:SDSS_lowCS}) and strong covariate shift (\ref{form:SDSS_strongCS}) scenarios on SDSS data, using a set of five predictors for propensity score estimation. The number of galaxies in source and target stratum, as well as the mean outcome (redshift $z$) are presented. 
\label{table:sdss_strata_comp_low-strong}}
\begin{tabular}{ l l r r  } 
\hline
\multirow{2}{*}{} &   &  weak covariate shift, 5 covariates    & strong covariate shift, 5 covariates  \\
Stratum & Set & \#galaxies (Mean $z$) & \#galaxies (Mean $z$) \\ 
\hline
\hline
\multirow{1}{*}{1} 
&  Source & 1468 (0.09) & 1603 (0.07) \\ 
&  Target & 145 (0.09) & 0 (NaN) \\  
\hline
\multirow{1}{*}{2} 
&  Source & 1220 (0.13) &  1579 (0.10) \\ 
&  Target & 602 (0.11) & 4 (0.10) \\ 
\hline
\multirow{1}{*}{3} 
&  Source & 740 (0.20) & 637 (0.24) \\ 
&  Target & 1275 (0.21) & 1494 (0.28) \\ 
\hline
\multirow{1}{*}{4} 
&  Source & 301 (0.37) &  73 (0.36) \\ 
&  Target & 1958 (0.37) & 2297 (0.33) \\ 
\hline
\multirow{1}{*}{5} 
&  Source & 271 (0.47) &  108 (0.41) \\ 
&  Target & 2020 (0.44) &  2205 (0.37) \\ 
\hline
\hline
\multirow{1}{*}{All} 
&  Source & 4000 (0.17) &  4000 (0.13) \\ 
&  Target & 6000 (0.33) &   6000 (0.34)\\
\end{tabular}
\end{table}

\paragraph{Summary -- Conditional Density Estimation:} Figure~\ref{figure:append_photoZ_results}  compares the resulting target risk $\hat{R}_T(\hat{f})$ across models and covariate sets, for the weak covariate shift (top row) and the strong covariate shift scenario (bottom row). 
These results, combined with those from the medium covariate shift scenario in Figure~\ref{figure:photoZ_results}, 
reinforce the advantage of using $\textit{StratLearn}$, especially for higher dimensional covariate sets.  
In Figures~\ref{figure:photoZ_results} and \ref{figure:append_photoZ_results}, 
$\textit{StratLearn}_\textit{Series}$ performs well in the setup with the fewest covariates (left column). In the weak covariate shift scenario, this leads to an additional boost in performance of $\textit{StratLearn}_\textit{Comb}$ (which combines $\textit{StratLearn}_\textit{ker-NN}$ and $\textit{StratLearn}_\textit{Series}$, following (\ref{formula:comb})). In the setups with higher dimensional covariates, the Series methods (using either \textit{StratLearn} or the weighting methods) exhibit relatively poor performance overall. In these cases, the $\textit{StratLearn}_\textit{Comb}$ predictions rely solely on  $\textit{StratLearn}_\textit{ker-NN},$ 
which again demonstrates the successful empirical risk minimization of (\ref{form:strata_risk}), inasmuch as $\textit{StratLearn}_\textit{Comb}$ automatically selects the better \textit{target} model (i.e., $\textit{StratLearn}_\textit{ker-NN}$) in each stratum, based on the respective empirical \textit{source} risk estimates in each stratum.
Overall, IPS and NN weighting are most competitive with \textit{StratLearn}, in addition to KLIEP in the low dimensional, strong covariate shift setup.
Note that the performances of  $\textit{IPS}_\textit{Comb}$,  $\textit{KLIEP}_\textit{Comb}$ and  $\textit{uLSIF}_\textit{Comb}$ are 
degraded in some cases (e.g., Figure~\ref{figure:append_photoZ_results}, upper left panel) with inclusion of the poorly performing $\textit{Series}$ predictions, illustrating that the weighted empirical \textit{source} risk minimization ((\ref{formula:SDSS_source_loss}), as a form of (\ref{form:risk_shimodaira})) does not lead to \textit{target} risk minimization in these situations.
With increasing number of covariates, none of the weighted models yield results that are  competitive with \textit{StratLearn}.

Table~\ref{table:sdss_strata_comp_medium_all}  presents the composition of the five \textit{StratLearn} strata for the medium covariate shift scenario (described in Section~\ref{section:photoZ}) for each of the three sets of covariates. Each source stratum contains a sufficient sample size to fit the conditional density estimators for prediction on the respective target stratum. The overall redshift averages in the source (0.11) and target data (0.28) are very unequal, a consequence of the covariate shift. Within strata, the redshift averages are well aligned between source and target, indicating improved balance after stratification.
Notably, the composition of the strata in the high-dimensional covariate setup is similar to the that of the lower-dimensional setups, with 
well-balanced strata. This is an indication of the robustness of \textit{StratLearn} with respect to high dimensional (noisy) sets of covariates.
Table~\ref{table:sdss_strata_comp_low-strong} compares the five \textit{StratLearn} strata for the weak and strong covariate shift scenarios, again with well-aligned average redshift within each stratum. Note that for the strong covariate shift scenario there are almost no galaxies in the first two target strata. Thus, effectively only data in source stratum 3 to 5 are used for target prediction in the respective strata.
These examples illustrate the advantage of using only a small subset of the source data when formulating predictions for individuals in the target, where the subset is chosen for its similarity to the target data in question. This is a markedly different strategy to the widespread practice of including all possible available data when fitting  machine learning models.

\bibliographystylesupp{JASA}
\bibliographysupp{./supp.bbl}

\end{document}